%% file: main.tex
\documentclass[10pt,twocolumn,letterpaper]{article}

\usepackage[pagenumbers]{cvpr} 

\input{preamble}

\definecolor{cvprblue}{rgb}{0.21,0.49,0.74}
\usepackage[pagebackref,breaklinks,colorlinks,citecolor=cvprblue]{hyperref}
\usepackage[linesnumbered,ruled,vlined]{algorithm2e} 
\usepackage{amsmath}
\SetKwInOut{Input}{Input}
\SetKwInOut{Output}{Output}
\makeatletter
\DeclareRobustCommand\onedot{\futurelet\@let@token\@onedot}
\def\@onedot{\ifx\@let@token.\else.\null\fi\xspace}
\def\eg{\emph{e.g}\onedot} 
\def\ie{\emph{i.e}\onedot}

\makeatother

\definecolor{redcolor}{RGB}{215,25,28}
\definecolor{violetcolor}{RGB}{239, 66, 245}

\newcommand{\ourmethod}{COMBO}
\newcommand{\ourmethodspace}{COMBO }

\def\paperID{15570} 
\def\confName{CVPR}
\def\confYear{2024}

\title{A Unified Framework for Connecting Noise Modeling to Boost Noise Detection}

\author{Siqi Wang \quad Chau Pham \quad Bryan A. Plummer \\
 Boston University \\
{\tt\small\{siqiwang, chaupham, bplum\}@bu.edu}
}

\begin{document}
\maketitle
\input{sec/1_abstract}    
\input{sec/2_intro}

\input{sec/3_related_work}

\input{sec/4_methods}
\input{sec/5_experiments}

\input{sec/6_conclusion}
{
    \small
    \bibliographystyle{ieeenat_fullname}
    \bibliography{main}
}

\input{sec/X_suppl}

\end{document}

%% file: preamble.tex
\usepackage[dvipsnames]{xcolor}


%% file: sec/1_abstract.tex
\begin{abstract}
Noisy labels can impair model performance, making the study of learning with noisy labels an important topic. Two conventional approaches are noise modeling and noise detection. However, these two methods are typically studied independently, and there has been limited work on their collaboration. In this work, we explore the integration of these two approaches, proposing an interconnected structure with three crucial blocks: noise modeling, source knowledge identification, and enhanced noise detection using noise source-knowledge-integration methods. This collaboration structure offers advantages such as discriminating hard negatives and preserving genuinely clean labels that might be suspiciously noisy.  Our experiments on four datasets, featuring three types of noise and different combinations of each block, demonstrate the efficacy of these components' collaboration.  Our collaborative structure methods achieve up to a 10\% increase in top-1 classification accuracy in synthesized noise datasets and 3-5\% in real-world noisy datasets. 
The results also suggest that these components make distinct contributions to overall performance across various noise scenarios. These findings provide valuable insights for designing noisy label learning methods customized for specific noise scenarios in the future. 
Our code is accessible to the public. \footnote{Available at \textbf{\url{https://github.com/SunnySiqi/COMBO}}}
\end{abstract}

%% file: sec/2_intro.tex
\section{Introduction}
\label{sec:intro}

\begin{figure}[t]
\centering
\includegraphics[width=\columnwidth]{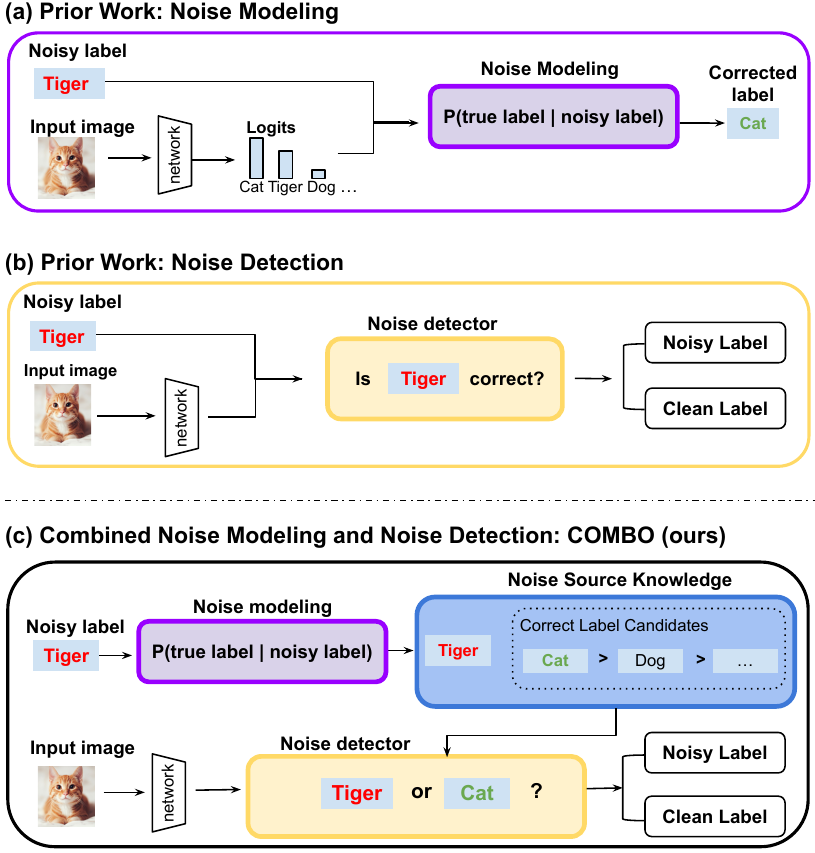}
\caption{\textbf{A Unified Framework for Connecting Noise Modeling.} 
In prior work, two distinct methodologies have been utilized for learning with noisy labels: noise modeling and noise detection. \textbf{(a) Noise modeling} involves using a noise transition matrix to align the training loss on noisy data with that of clean data, thereby predicting true labels while training with noisy labels; \textbf{(b) noise detection} concentrates on differentiating between clean and noisy samples, and subsequently training the network using the identified clean labels; \textbf{(c) \ourmethod} (ours) leverages the noise source information inferred from the noise transition matrix, commonly used in noise modeling, to augment the effectiveness of noise detection methods. 
}
\label{Fig:Connection}
\end{figure}

Automated data collection methods have led to increased dataset generation with minimal human effort. However, these techniques often introduce noise, particularly in label data from sources like web crawling~\cite{krishna2016embracing, yan2014learning}, which is costly and challenging to deal with. Learning with Noisy Labels (LNL)~\cite{natarajan2013LNL} offers an effective and efficient alternative. It allows the training of models with strong resistance to noise, resulting in enhanced generalization and mitigating the adverse effects of noisy labels~\cite{arpit2017closer}. Prior work can be separated into two main approaches: \textit{Noise modeling} and \textit{noise detection},
as shown in Fig.~\ref{Fig:Connection}. Noise modeling methods (Fig.~\ref{Fig:Connection}a)  train the consistent classifier with estimated noise transitions~\cite{scott2015rate, liu2015classification, yao2020dual, xia2019anchor, zhang2021learning, kye2022learning, cheng2022instance}. In contrast, noise detection methods (Fig.~\ref{Fig:Connection}b) estimate the probability of the label being clean for each sample, and employ different training strategies for clean and noisy samples~\cite{kim2021fine, crust, wei2022self, liu2020ELR,iscen2022ncr, han2018co-teaching,karim2022unicon, li2020dividemix}. Noise modeling methods boast robust theoretical foundations, whereas detection methods showcase state-of-the-art (SOTA) performance across numerous benchmark datasets~\cite{yao2023better}. However, the distinct objectives of aligning the classifier with either clean or noisy labels in these two primary approaches have traditionally led to isolated research, with almost no intermediary framework to bridge them together.


To this end, we explore a unified framework for Connecting nOise Modeling to Boost nOise detection (\textbf{\ourmethod}). The core of this approach is the utilization of noise source knowledge as the bridge to establish connections between the two themes of prior work (see Fig.~\ref{Fig:Connection}c for the general pipeline.) \ourmethodspace comprises three key components: Noise Modeling, Noise Source Identification, and Noisy Label Detection \& Training. 
Contrasted with the noise detection method illustrated in Fig.~\ref{Fig:Connection}b, the central question in detection shifts from ``\textit{Is this label correct?}" to ``\textit{Which label is correct, A or B}?" 

This revised inquiry not only adds specificity but also yields twofold advantages. First, detecting samples with relative similarity scores is more reliable than searching for a fixed similarity threshold, especially for distinguishing similar features with noisy labels. For example, in Fig.~\ref{Fig:Connection}b, a cat and a tiger might share similar features, making it possible to fool the model ``tiger" is a clean label of a cat. However, when introducing another class for comparison, \ie, given both ``tiger" and ``cat" as choices, the cat image has relatively higher similarity score with cats than tigers, thus it is easier for the model to detect ``tiger" as a noisy label.

The second advantage is that comparisons also help preserve less probable clean samples. Specifically, if the probability of a clean label is low but exceeds that of the noise source, the sample is still retained as a clean sample for training. For example, consider a cat image with a wooded background resembling a bird image in a similar setting. The likelihood of the cat image bearing a clean cat label might be low, potentially categorizing it as a noisy label in Fig.~\ref{Fig:Connection}b. However, since the bird is not regarded as the noise source for the cat (likely the dog is), the model deduces that the image is more ``likely" to be a cat rather than a dog. Consequently, this bird-like cat is still identified as clean in our methods.

The work most similar to ours is LNL+K~\cite{wang2023lnl+}, which suggests incorporating noise source knowledge in noise detection methods. However, the knowledge-integrated methods proposed in their work require the additional input of noise knowledge and it does not offer general guidelines on how to learn noise sources from noise estimations. Our framework, \ourmethod, moves beyond the prior knowledge assumption limit by proposing a general way of obtaining noise source knowledge from noise modeling methods. The integration of these components not only enhances the performance of individual blocks but also allows for a comprehensive understanding of their distinct contributions to overall performance across various noise settings.

To demonstrate \ourmethod's effectiveness, we conducted extensive experiments using four noise modeling methods~\cite{yao2020dual, zhang2021learningtvd, yong2022holistic}, including the baseline method \textit{growing cluster} we adjusted, two knowledge-integrated noise detection methods~\cite{kim2021fine, karim2022unicon, wang2023lnl+}, and two training methods: \textit{sample selection} and \textit{train with pseudo labels}. Comparing various combinations highlighted distinct contributions of each \ourmethod's component across diverse noise scenarios on both synthesized and real-world datasets. 
To gain a deeper understanding of the improvements in noise detection methods, we analyze the correlations between sample selection scores and classification accuracy.
Intriguingly, in low noise settings, sample selection recall strongly influenced classification accuracy, while high noise ratios showed a difference in classification accuracy for a comparable F1 selection score. 

To summarize, our contributions are:
\begin{itemize}
\setlength\itemsep{0em}
    \item We introduce \ourmethod, an innovative framework that establishes effective collaboration between noise modeling and detection for LNL tasks. 
    \item We analyze how the components within \ourmethodspace yield distinct contributions in varying noise settings, offering valuable insights for designing specific models tailored to particular noise scenarios.
    \item Specific combinations of methods within the \ourmethodspace structure enhance performance by up to 10\% in some noise settings and by an average of 4\% across the synthesized datasets. Additionally, these combinations yield a performance gain of 3\% to 5\% on real-world datasets.

\end{itemize}

%% file: sec/3_related_work.tex
\section{Related Work}

COMBO framework connects two important branches of LNL methods: noise modeling and noise detection.

\smallskip
\noindent\textbf{\textit{Noise Modeling}} methods aim to align models trained on noisy data with the optimal classifiers that minimize errors in clean data~\cite{yao2023better}. This approach trains the model to predict the clean label and utilize the noise transition matrix $T(X = x)$, where $T_{ij}(X = x) = P(\widetilde{Y}= j \mid Y = i, X = x)$ ($Y$ is the clean label, $\widetilde{Y}$ is the noisy label), to transfer the label to the corresponding ``noisy" label, so that making use of the noisy label for backpropagation and optimize the network. This process relies on estimating the transition matrix. Some studies introduced the concept of ``anchor points" — data points demonstrating high probabilities of belonging to specific classes~\citep{scott2015rate, liu2015classification, menon2015learning, patrini2017making}. While clean anchor points are not always available in the dataset, more recent research endeavors to estimate the transition matrix using only noisy data~\citep{li2021provably, yao2020dual, xia2019anchor, zhang2021learning, kye2022learning, cheng2022instance}. Their estimation is either with statistic methods, matrix decomposition~\cite{yao2020dual, patrini2017making, liu2015classification} or train a network to predict the elements of the transition matrix~\cite{yong2022holistic,yang2022estimating}.  
These methods train on the complete noisy dataset, with a strong dependence on the accuracy of the noise transition matrix. This poses a considerable challenge, particularly in scenarios with higher noise ratios. 

\smallskip
\noindent\textbf{\textit{Noise Detection}} approaches aim to distinguish between clean and noisy labels. To identify these labels, two primary detection methods are employed: \textit{loss-based} methods, detecting noisy samples by their high losses~\citep{jiang2018mentornet, li2020dividemix, arazo2019unsupervised_loss}, and \textit{probability distribution-based} strategies that identify clean samples with high confidence~\citep{hu2021p, torkzadehmahani2022confidence, nguyen2019self, tanaka2018joint, li2022selective}. However, these methods fall short with hard negative samples such as samples along distribution boundaries. Consequently, samples selected by these means are more likely to be perceived as ``easy" rather than genuinely ``clean." There are also feature-based methodologies proposed that leverage high-dimensional features before the softmax layer, which are less influenced by noisy labels~\citep{crust, kim2021fine} and have been shown to be effective in handling noisy labels~\citep{li2020lastlayeroverfitting, yao2020searching, bai2021understanding}. To distinguish between training noisy and clean samples, various techniques have been employed: adjusting loss functions~\citep{wei2022self,ma2020normalized,iscen2022ncr,xu2019l_dmi, zhang2018generalized}, utilizing regularization methods~\citep{liu2020ELR,xia2021param, hu2019simple, liu2022robust}, iterative learning with selected clean samples in multiple rounds~\citep{cordeiro2023longremix,shen2019learning_iterative, wu2020topological}, and employing semi-supervised learning (SSL) for training noisy samples~\citep{sohn2020fixmatch, tarvainen2017mean, li2020dividemix, karim2022unicon}. Notably, many statistically inconsistent methods often disregard the valuable resource of noise distribution knowledge within the context of LNL. In recent work, LNL+K offers a distinctive contribution by leveraging noise source knowledge to identify clean samples within these methodologies~\cite{wang2023lnl+}. Although LNL+K exhibits promise in bridging the divide between classifier-consistent and classifier-inconsistent methods, it does not offer a complete framework for learning noise sources from noise estimates. Our work addresses this gap by utilizing noise source knowledge to forge essential connections.
\smallskip

%% file: sec/4_methods.tex
\section{Connecting nOise Modeling to Boost nOise detection
(\ourmethod)}
\label{sec:COMBO_methods}
\begin{figure*}[t]
\centering
\includegraphics[width=\textwidth]{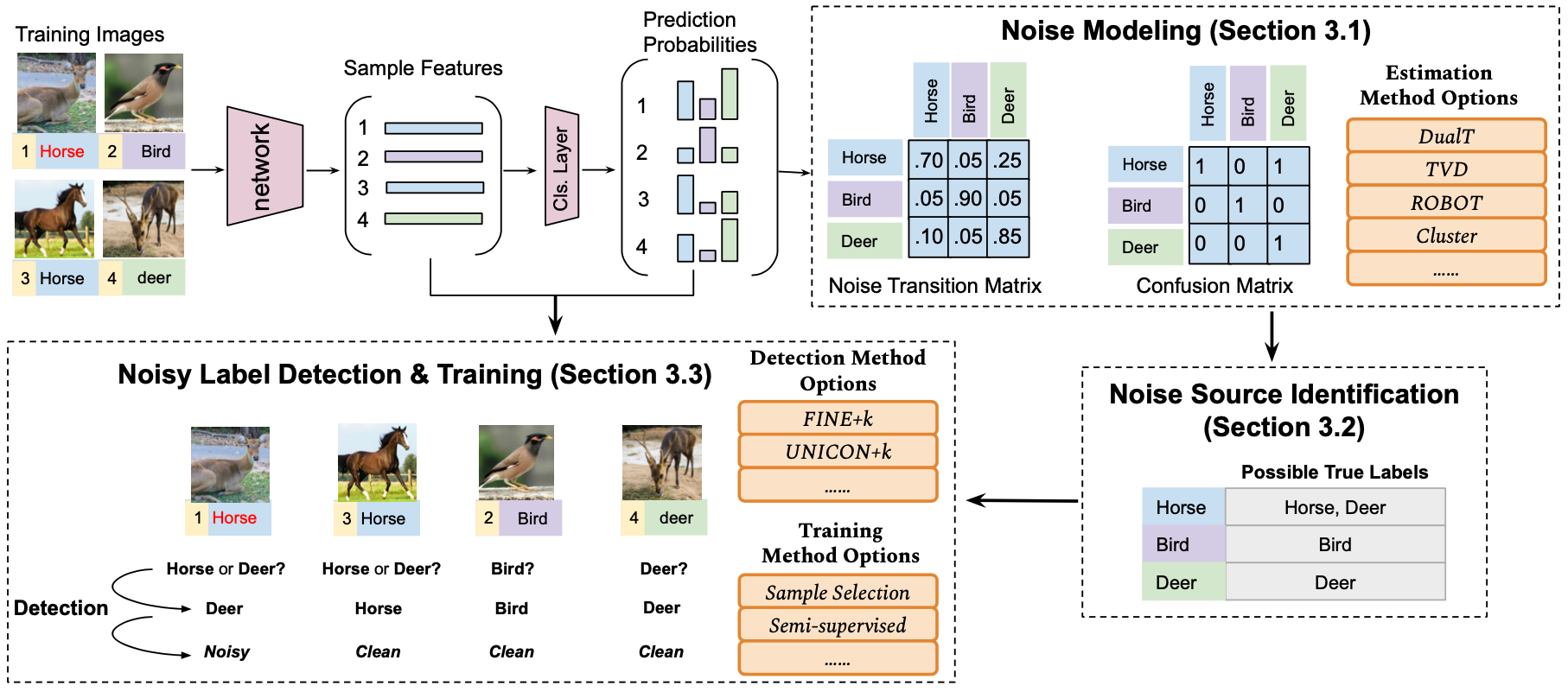}
\caption{\textbf{\ourmethodspace updates the training set.} The process begins by generating feature vectors from the network and prediction probabilities from the classifier layer. These, along with dataset labels are used in the \textit{Noise Modeling} section, producing a noise transition matrix or confusion matrix depending on the chosen method. Subsequently, in the \textit{Noise Source Identification}, sources of noise are determined from the matrix. These identified noise sources, along with the features, labels, and predictions, serve as input for knowledge-integrated \textit{Noisy Label Detection \& Training} methods. The detection methods distinguish between clean and noisy samples and then employ various training strategies, such as sample selection or semi-supervised techniques with pseudo labels. After these three interactive components, the data loader is updated, and the network continues training with the newly updated dataset in the subsequent epoch.  It is worth noting that the frequency of dataloader updates is determined by a hyperparameter, and each update begins with the complete training set. }
\label{Fig:Overall}
\end{figure*}
In this section, we introduce our \ourmethodspace approach which consists of three components: \textit{noise estimation}, \textit{noise source identification}, and \textit{noisy label detection \& training}, as shown in Fig.~\ref{Fig:Overall}.
We provide formal definitions for each component, followed by detailed explanations in subsequent subsections.


Let's consider a dataset $D =\{(x_i, \widetilde{y_i})\}_{i=1}^n$ with $n$ samples. $\widetilde{y_i} \in K$, where $K =\{1, 2, ..., k\}$ represents the categorical label across $k$ classes. Labels $\{\widetilde{y_i}\}_{i=1}^n$ may contain noise, while the true labels $\{y_i\}_{i=1}^n$ remain unknown. The input feature for the final classifier layer of sample $x_i$ is denoted as $f(x_i)$, and the classifier output is a  probabilistic distribution across all classes as $p(x_i) \in \mathbb{R}^k$, where $p(x_i, j)$ refers to the probability of sample $x_i$ with label $j$, the prediction is $\hat{y_i} = \arg\max_{j} p(x_i, j)$. 

The noise modeling module takes the feature $f(x)$, label $\widetilde{y}$, and prediction distributions $p(x)$ as the input, and the output is either a $k \times k$ transition matrix $T_m$ or a $k \times k$ confusion matrix $C_m$. In the transition matrix $T_m$, the element $T_m[i][j]$ represents the probability $P(\widetilde{y}=j \mid y=i)$, marking the likelihood of observing the noisy label $j$ given the true label $i$. In the confusion matrix $C_m$, the entry $C_m[i][j]$ denotes the count $|\{x \mid \widetilde{y_i}=i \wedge \hat{y_i} = j\}|$, indicating the number of samples with label $i$ and prediction $j$. 

Utilizing $T_m$ or $C_m$ as the input, The noise source identification module outputs a dictionary \textit{NS}, where \textit{NS}$[c]$ presents the potential true labels for class $c$.
Together with the feature $f(x)$, label $\widetilde{y}$, prediction $p(x)$, and the noise source \textit{NS}, the noise detection module outputs the probability of the label being clean, denoted as $P_{clean}(x_i) = P (y_i=\widetilde{y_i})$.

\subsection{Noise Modeling}
Noise Modeling aims to extract the noise pattern from the network output, specifically from the sample features and their predicted probability predictions. The noise transition matrix represents one form of this pattern, displaying the transition probability from true labels to noisy labels. Another form is the confusion matrix between predictions and noisy labels. The classes that confuse the model the most are likely to be each other's noise sources. In simpler terms, incorrect predictions can either be due to the model's errors or incorrect labels. It is important to note that while noise sources can be inferred from both the noise transition matrix and confusion matrix, these matrices convey distinct meanings, and the confusion matrix cannot serve as a replacement for the transition matrix in the LNL consistent-classifier methods.

We explore a growing cluster method for estimating the noise pattern using the confusion matrix and list a few options of LNL consistent-classifier methods~\cite{yao2020dual, zhang2021learningtvd, yong2022holistic} that can be employed in \ourmethodspace framework.

\subsubsection{Estimation methods}
\label{sec:estimation methods}
\smallbreak
\noindent\textbf{Growing cluster for noise modeling.}~\cite{kaufman1987clustering} Inspired by the observation that high-dimensional features are less susceptible to noisy labels~\cite{li2020lastlayeroverfitting, yao2020searching, bai2021understanding}, we explore growing clusters in the feature space and apply this cluster label to the confusion matrix. The cluster generation involves two steps: initializing clusters using confident, prediction-label consistent samples, and then expanding these clusters based on descending confidence scores of sample predictions.

Cluster anchor points for a specific class $k$, denoted as $C_k^{\text{\textit{init}}}$, are outlined as:

\begin{equation}
S_1 = \{x_i  \mid \widetilde{y_i} = k \wedge \hat{y_i} = k \wedge p(x_i, k) > \alpha\}
\label{eq:cluster_s1}
\end{equation}

\begin{equation}
S_2 = \{x_i  \mid \widetilde{y_i} = k \wedge p(x_i, \hat{y_i}) < \beta\}
\label{eq:cluster_s2}
\end{equation}

\begin{equation}
C_k^{\text{\textit{init}}} = 
\begin{cases}
   S_1, & \text{if } S_1 \neq \varnothing \\
   S_2, & \text{otherwise}
\end{cases}
 \label{eq:cluster_anchor}
\end{equation}

 \noindent where $\alpha$, $\beta$ represent the selection thresholds, which could be percentile values derived from $p(x_i, k)$. The intuition behind these equations is as follows: if prediction-label consistent samples exist, the anchors are those with high probabilities among these samples, as shown in Eq.~\ref{eq:cluster_s1}.  However, if there are no such samples for class $k$, then according to Eq.~\ref{eq:cluster_s2}, samples with noisy label $k$ and low probabilities in the prediction class are chosen. 
 In other words, if none of the samples identified with label $k$ predict that class, the initial cluster points considered reliable are those that demonstrate the lowest confidence in their ``incorrect" predictions, or those that exhibit the greatest uncertainty in predictions for other classes.
 The remaining unselected samples are then sorted based on the descending order of $p(x_i, \hat{y_i})$ values regardless of the predictions $\hat{y_i}$. This means that samples with higher confidence in their predictions are added to the cluster earlier. The centroid of each cluster, denoted as $C_k^{\text{\textit{*}}}$, is computed as the mean feature vector of the samples:

\begin{equation}
C_k^{\text{\textit{*}}} = \frac{\sum_{x_i \in \text{C}_{k}} f(x_i)}{|\text{C}_{k}|}
\label{eq:cluster_centroid}
\end{equation}

\noindent These centroids are dynamically updated each time a new point is added to the corresponding cluster by Eq.~\ref{eq:cluster_centroid}.
\smallbreak
\noindent\textbf{DualT}~\cite{yao2020dual} estimates the noise transition matrix in a divide-and-conquer paradigm. It introduces an intermediate class and factorizes the estimation of a single transition matrix into two easy-to-estimate matrices. In the experiment, we set the intermediate class as the model prediction class. Thus the noise transition matrix $T_m[i][j] = P(\widetilde{Y}=j \mid Y=i) = \sum_{l \in K} P(\widetilde{Y}=j \mid \hat{Y}=l, Y=i) \cdot
 P(\hat{Y}=l \mid Y=i)$, which can be simplified as $T_m[i][j] = P(\widetilde{Y}=j \mid Y=i) = \sum_{l \in K} P(\widetilde{Y}=j \mid \hat{Y}=l) \cdot P(\hat{Y}=l \mid Y=i)$ shown in their proof. $P(\widetilde{Y}=j \mid \hat{Y}=l)$ can be estimated from the direct count in the predictions with no error. For $P(\hat{Y}=l \mid Y=i)$, if the anchor points (samples that are confident to have clean labels) exist, then no estimation for this as well, otherwise it will apply the traditional T-estimator on this part.  
 
\smallbreak
\noindent\textbf{Total Variation Regularization} (TVD)~\cite{zhang2021learningtvd} estimates the noise transition matrix and trains a classifier simultaneously. A Dirichlet posterior is applied for the noise transition matrix, whose concentration parameters are updated with a confusion matrix from the classifier. With the observation that the ``cleanest" class-posterior has the highest pairwise total variation, they regularize the predicted probabilities more distinct to find the optimal estimation.

\smallskip
\noindent\textbf{RObust Bilevel OpTimzation} (ROBOT)~\cite{yong2022holistic}
addresses the challenge of estimating the noise and training the classifier together by splitting it into ``inner" and ``outer" problems, with separate networks trained for each problem. In the inner loop, the transition matrix $T_m$ is fixed, and the classifier is optimized. Vice versa, in the outer loop, the classifier parameters are fixed and the transition matrix $T_m$ is updated.

\subsection{Noise Source Identification}
The objective of this component is to summarize the noise sources \textit{NS} from the output of the noise modeling $T_m$ or $C_m$. 
For both the confusion matrix $C_m$ and the transpose of the noise transition matrix $T_m^T$, the rows represent noisy labels while the columns indicate predicted or true labels. Each column shows the distribution of noisy labels within that specific class. Rows with notably high values (excluding diagonal values) in a particular column are more likely to contain noise. For example, considering column 1 in $C_m$ representing samples in the class ``1," a significantly large number of samples labeled as ``9" suggests that these ``9" labeled samples share more feature similarities with class ``1." Consequently, ``1" is added as the noise source for ``9", $\text{\textit{NS}}[9] = 1$ implying that for samples labeled as ``9," the label might be noisy, while the true label could be ``1."

Humans might easily identify potential noise sources by observing classes with notably high transition probabilities or a considerable number of mispredictions from the estimation matrices. However, defining the threshold as the ``significant" number is challenging for the model. To address this, we employ the Gaussian Mixture Model (GMM) on each column to check the ``peaks" in the distribution.  Bayesian information criterion (BIC) values guide us in determining how well the number of distributions (peaks) fits the distribution. We test with two or three peaks: two correspond to ``true-label-predictions/transitions" and ``minor errors," while three include an extra group for ``noisy labels." Classes exhibiting higher probabilities in this ``noisy labels" group are more likely to contain noise.

\subsection{Noisy Label Detection and Training}
With the noise source knowledge \textit{NS} from the previous identification step, we apply LNL+K methods in this section for noise detection~\cite{wang2023lnl+}. LNL+K methods are the adaptions of LNL methods with an additional input of the noise source knowledge. Most LNL methods select the samples with maximum similarities within the class as clean samples, while LNL+K methods add additional comparisons to the noise source classes, so the clean samples should also have the least similarities with the noise source class. 

\subsubsection{Noisy Label Detection method choices}
\smallbreak
\noindent\textbf{FINE}~\cite{kim2021fine} and \textbf{FINE$^{+k}$}~\cite{wang2023lnl+}
use feature eigenvectors for detection. The alignment between a sample and its label class is determined by the cosine distance between the sample’s features and the eigenvector of the class feature gram matrix, which serves as the feature representation of that category. FINE then fits a Gaussian Mixture Model (GMM) on the alignment distribution to divide samples into clean and noisy groups - the clean group has a larger mean value, which refers to a better alignment with the category feature representation. The knowledge-integrated method FINE$^{+k}$ makes adaptations on the clean probability by adding the difference between label-class and noise-source-class alignment. Clean samples have higher alignment differences, while noisy labels have lower values.
\smallbreak
\noindent\textbf{UNICON} ~\cite{karim2022unicon} and \textbf{UNICON$^{+k}$}~\cite{wang2023lnl+}
estimate the clean probability with Jensen-Shannon divergence (JSD), a metric for distribution dissimilarity. Disagreement between predicted and one-hot label distributions is utilized, ranging from 0 to 1, with smaller values indicating a higher probability of a clean label. UNICON$^{+k}$ integrates the noise source knowledge by adding the comparison of JSD with the noise source class. If the sample’s predicted distribution aligns more with the noise source, it is considered noisy.

\subsubsection{Noise Training method choices}
After dividing the dataset into clean and noisy samples, distinct training approaches are applied. One method exclusively involves training with the selected clean samples, while the other applies semi-supervised techniques to the noisy samples.
\smallbreak
\noindent
\textbf{Sample selection}~\cite{crust, kim2021fine} discards the noisy samples and trains only with selected clean samples. 
\smallbreak
\noindent
\textbf{Semi-supervised learning (SSL)}~\cite{learning2006semi, karim2022unicon}  uses pseudo-labels for noisy samples. In our experiments, cluster labels learned in the estimation section are utilized as pseudo-labels. Mix-up~\cite{zhang2017mixup} augmentation is also applied between the samples from clean and noisy groups.

%% file: sec/5_experiments.tex
\section{Experiments}
\subsection{Baselines} 
We include three groups of baselines: Standard, LNL Regularization,  and Oracle training. Experiments are designed both on these single methods and the combination under our \ourmethodspace framework. 

\smallbreak
\noindent \textbf{Regularization methods.}
Sparse over-parameterization (SOP)~\cite{liu2022robust} is the method of exploiting algorithmic regularization to recover and separate the corruptions based on introducing more parameters to an already overparameterized model. It uses an extra variable $s_{*i}$ to represent the difference between the observed label $y_i$ and its corresponding clean label, and the goal is to minimize the discrepancy between $\hat{y_i} + s_{*i}$ and $y_i$.

\smallbreak
\noindent
\textbf{Standard and Oracle learning on noisy labels.}
Standard training is the training of the entire dataset without any LNL methods. Oracle training is the training of only the clean dataset, which is available for synthesized datasets.

\subsection{Datasets and Metrics}
\subsubsection{CIFAR Dataset with Synthesized Noise}
CIFAR-10/CIFAR-100~\citep{krizhevsky2009cifar} dataset contains 10/100 classes, with 5000/500 images per class for training and 1000/100 images per class for testing. We applied two synthesized noises with two noise ratios 20\% and 40\%. Noting here, the noise ratio specified in the experiment is for classes containing the noise, not the overall noise ratio.
\smallbreak
\noindent\textbf{Pairwise Noise} is a bi-directional noise. The noise source for the noisy label is only one, but it is mutual. Pairwise noise can be used to simulate the confusing class pairs in the real-world dataset. In the synthesized noise experiments, labels are corrupted to their visually similar classes, \eg, horse and deer. See the appendix for the complete noise pairs in our experiments.

\smallbreak
\noindent\textbf{Dominant Noise}  represents a one-directional noise pattern, and its challenge lies in the potentially large number of noise sources. LNL+K introduced this noise type in their work~\cite{wang2023lnl+}, and we follow the precise label corruption rules outlined in their research. It is worth noting that, while this noise type may seem similar to symmetric noise, the key distinction is that, even with the same noise ratio, the proportion of clean samples to the noisy samples attributed to each noise source is higher in this dominant noise setting.

\subsubsection{Real-World Datasets with Natural Noise}
\smallbreak
\noindent\textbf{Cell Datasets CHAMMI-CP}~\cite{ChenCHAMMI2023}
contains U2OS cell images from the Cell Painting datasets~\cite{bray2016cell}, used for comprehensive treatment screens involving chemical and genetic perturbations. CHAMMI-CP\footnote{Available at \textbf{\url{https://zenodo.org/record/7988357}}} sampled 7 compounds out of 1500, including ``control" (\ie, doing-nothing) group. The training data is five-channel cell morphology images and labels are treatments on the cells. However, certain cell samples labeled as treatments exhibit similar features to the control group. In our model aimed at identifying treatment effects, these samples with treatment labels but resembling the control group in features are considered noisy labels. For CHAMMI-CP, three treatments exclusive to the test set were removed, resulting in four classes: ``weak," ``medium," and ``strong" treatments, along with the ``control" class. This dataset encompasses 36,360 training images, 3,636 validation images, and 13,065 test images. 
\smallbreak
\noindent\textbf{Animal10N}~\cite{song2019selfie}
comprises human-labeled online images with noise. Acquired through predefined labels used as search keywords on multiple search engines, these images underwent human classification for the clean test set. The dataset includes 5 pairs of easily confusable animals, totaling 50,000 training images and 5000 testing images.


\subsection{Results}
\label{sec:results}

\begin{figure*}[t]
\centering
\begin{subfigure}[t]{\columnwidth}
    \centering
    \includegraphics[width=0.92\textwidth]{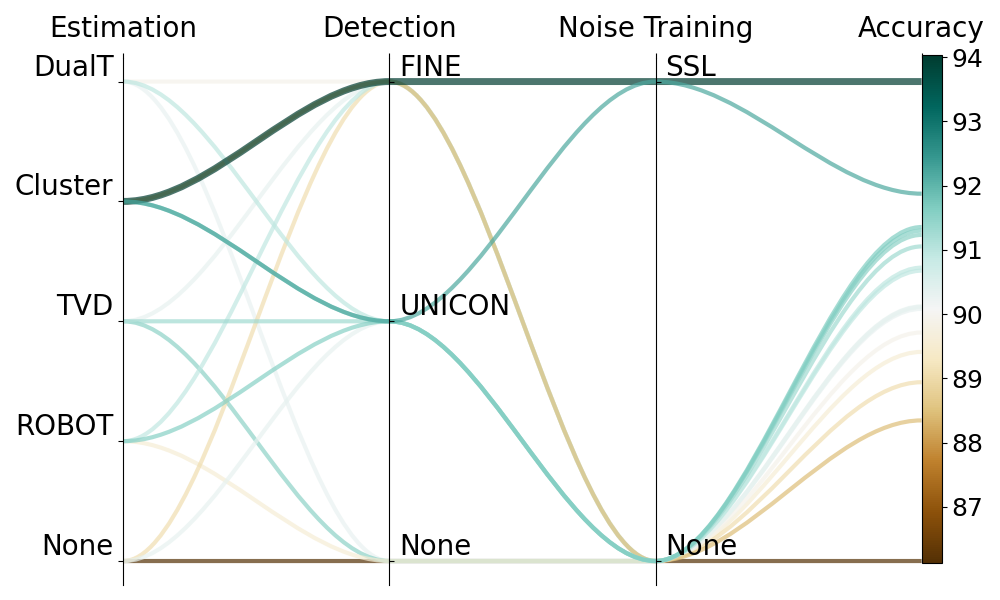}
    \caption{CIFAR-10 20\% pairwise noise}
    \label{fig:cifar10_asym02}
\end{subfigure}
\begin{subfigure}[t]{\columnwidth}
    \centering
    \includegraphics[width=0.92\textwidth]{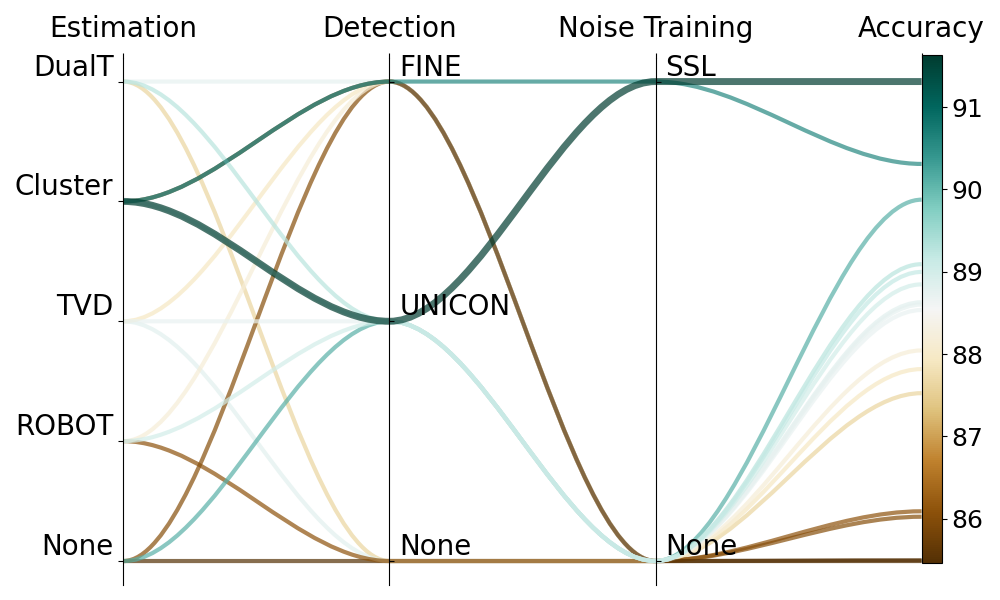}
    \caption{CIFAR-10 20\% dominant noise}
    \label{fig:cifar10_dominant02}
\end{subfigure}
\begin{subfigure}[t]{\columnwidth}
    \centering
    \includegraphics[width=0.92\textwidth]{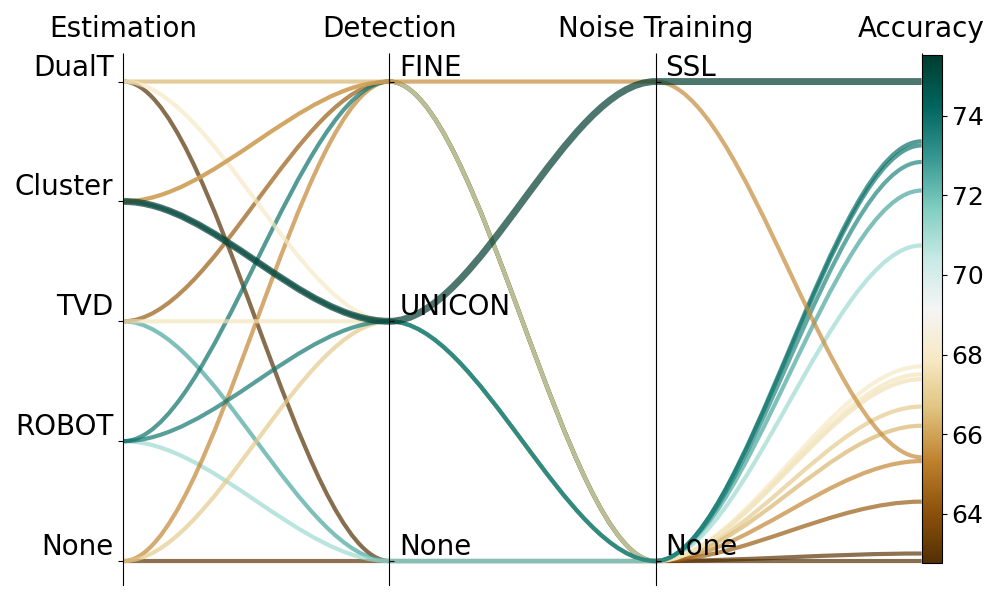}
    \caption{CIFAR-100 20\% pairwise noise}
    \label{fig:cifar100_asym02}
\end{subfigure}
\begin{subfigure}[t]{\columnwidth}
    \centering
    \includegraphics[width=0.92\textwidth]{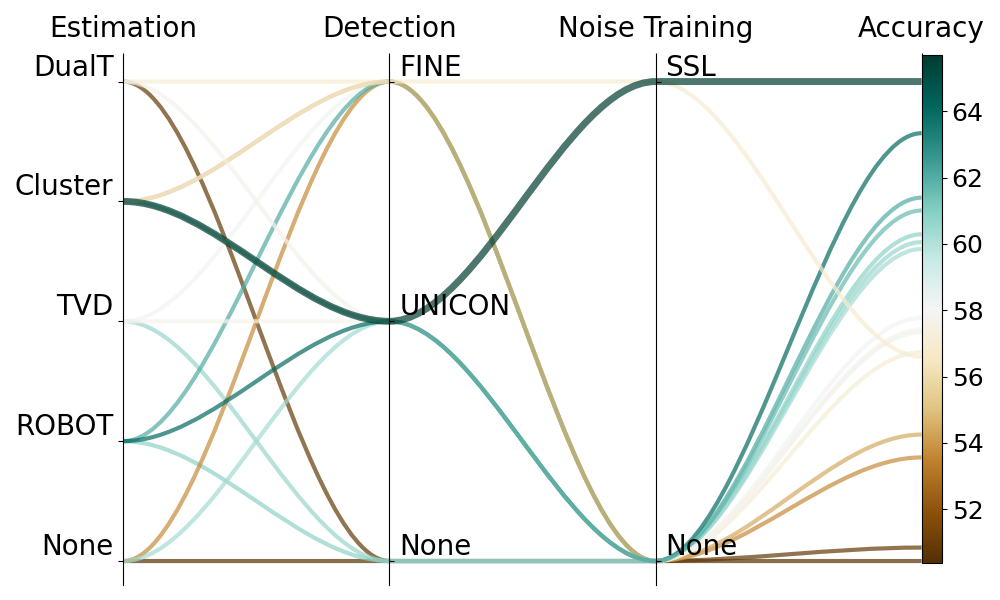}
    \caption{CIFAR-100 20\% dominant noise}
    \label{fig:cifar100_dominant02}
\end{subfigure}

\caption{\textbf{Component interactions in \ourmethodspace framework.} Impact of integrating different methods in three components under \ourmethod: \textit{noise estimation}, \textit{noise detection}, and \textit{noise training}. Each line in the plot represents a different combination of the components. We can trace the most effective strategies by tracking the lines that reach the highest point on the accuracy axis (rightmost column). In general, the absence of one or some components (denoted as \textit{None}) leads to lower accuracy, suggesting that combining these components is crucial in handling noisy data. Notably, using Semi-supervised training (\textit{SSL}) consistently achieves the best performance. See Sec.~\ref{result:combination} for discussion.}
\label{fig:parallel_charts}
\end{figure*}

\begin{figure*}[t]
\centering
\includegraphics[width=\textwidth]{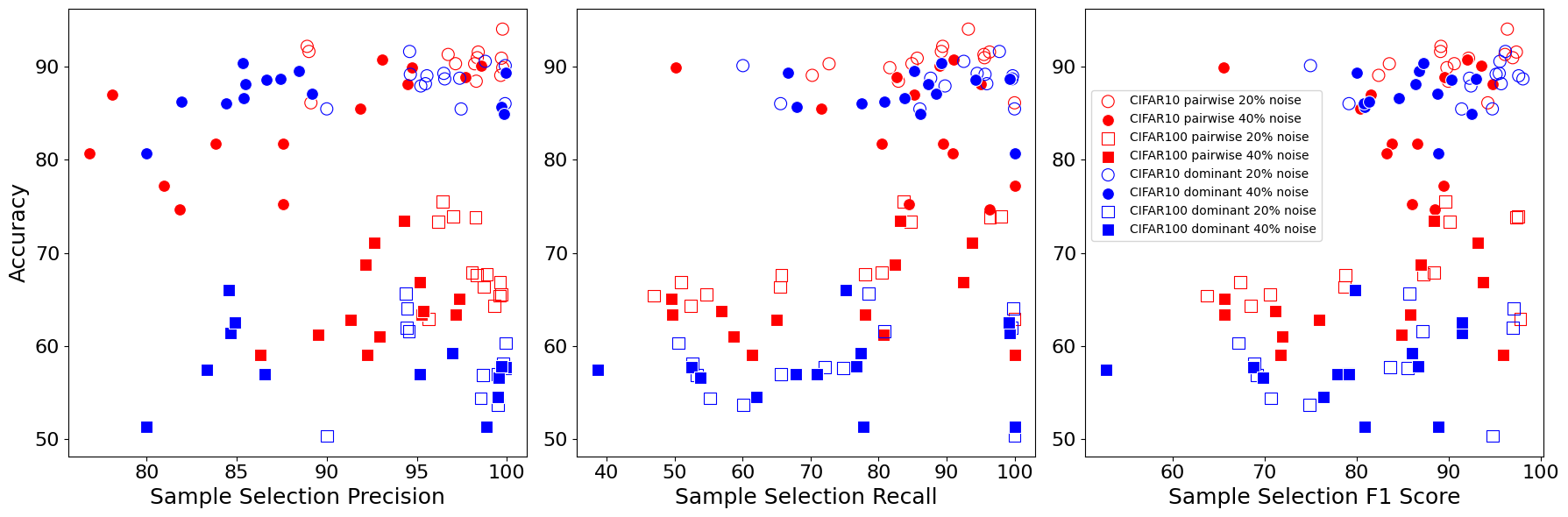}
\caption{\textbf{Relationship between Precision, Recall, F1 Scores of Sample Section and Classification Accuracy}. Each noise setting is characterized by three attributes: noise types represented by color (red for pairwise noise, blue for dominant noise), dataset types indicated by shape (circle for CIFAR-10, square for CIFAR-100), and noise ratios represented by fill-color (empty for 20\% noise, full for 40\% noise). Notably, a strong correlation exists between the sample selection F1 score and accuracy in less complex datasets like CIFAR-10, especially in settings with lower noise ratios. However, this correlation diminishes in more complex noise scenarios. See Section~\ref{result:F1-accuracy} for discussion.}
\label{Fig:precision_recall_f1_acc}
\end{figure*}

\begin{table*}[t]
    \centering
    \setlength{\tabcolsep}{5.5pt}
    \begin{tabular}{rlcccccccc}
    \hline
    & & \multicolumn{4}{c}{Pairwise Noise} & \multicolumn{4}{c}{Dominant Noise}\\
    & & \multicolumn{2}{c}{CIFAR-10} & \multicolumn{2}{c}{CIFAR-100} & \multicolumn{2}{|c}{CIFAR-10} & \multicolumn{2}{c}{CIFAR-100}\\
    \cline{3-10}
    & Noise Ratio & 20\%  & 40\% & 20\%  & 40\% & 20\% & 40\% & 20\% & 40\% \\
    \hline

    \textbf{(a)} & Standard training   & 86.12  & 77.18  & 62.96   & 59.07 & 85.47 & 80.68 & 50.37 & 51.40\\
        \hline
        \textbf{(b)} & SOP~\cite{liu2022robust}   & 92.85 & 89.93 & 72.60 & 70.58 & 89.86 & 89.30 & 62.47 & 61.33\\
        \hline
        \textbf{(c)} & DualT~\cite{yao2020dual}   & 90.28   & 75.35  & 62.76 & 57.74 & 87.62 & 87.54 & 50.80 & 50.76\\
        & TVD~\cite{zhang2021learningtvd}    & 91.50  & 83.20  & 72.62  & 65.48 & 88.79 & 88.12 & 60.56 & 57.19\\
        & ROBOT~\cite{yong2022holistic}    & 89.57  & 79.34  & 71.16  & 68.44 & 86.10 & 84.15 & 60.81 & 58.56\\
        \hline 
        \textbf{(d)} & FINE~\cite{kim2021fine}   & 89.07   & 85.51  & 65.42 & 65.11 & 86.03  & 85.70 & 53.68 & 51.36\\
        & UNICON~\cite{karim2022unicon}  & 90.31   & 89.91 & 66.87 & 63.43 & 90.11 & 89.39 & 60.34 & 57.74 \\
        \hline
        \textbf{(e)} & DualT~\cite{yao2020dual} + FINE$^{k}$~\cite{kim2021fine, wang2023lnl+}   & 89.89  & 88.87  & 66.36  & 62.80 & 88.76 & 87.09 & 57.04 & 54.53 \\
        & TVD~\cite{zhang2021learningtvd} + FINE$^{k}$~\cite{kim2021fine, wang2023lnl+} & 90.32   & 81.71 & 64.34 & 59.05 & 87.93 & 86.04 & 58.14 & 56.59\\
        & ROBOT~\cite{yong2022holistic} + FINE$^{k}$~\cite{kim2021fine, wang2023lnl+}   & 90.91   & 90.07   & 73.93   & 71.11 & 88.17 & 86.60 & 61.98 & 61.43\\
        & Cluster + FINE$^{k}$~\cite{kim2021fine, wang2023lnl+}   & 88.44  & 81.74  & 67.60 & 61.04 & 85.46 & 84.96 & 54.41 & 57.04\\
        & DualT~\cite{yao2020dual} + UNICON$^{k}$~\cite{karim2022unicon, wang2023lnl+}  & 90.96   & 75.28   & 67.94   & 63.42 & 89.28 & 89.53 & 57.68 & 57.80\\
        & TVD~\cite{zhang2021learningtvd} + UNICON$^{k}$~\cite{karim2022unicon, wang2023lnl+}   & 91.31  & 74.71  & 67.73 & 61.23 & 88.69 & 88.60 & 57.76 & 59.23\\
        & ROBOT~\cite{yong2022holistic} + UNICON$^{k}$~\cite{karim2022unicon, wang2023lnl+}   & 91.57   &88.09   & 73.82   & 66.89 & 89.02 & 86.29 & 64.04 & 62.56\\
        & Cluster + UNICON$^{k}$~\cite{karim2022unicon, wang2023lnl+}    & 91.63    & 87.01  & 73.38   & 68.71 & 89.18 & 88.09 & 61.57 & 57.44\\
     \hline
        \textbf{(f)} & Cluster + FINE$^{k}$~\cite{kim2021fine, wang2023lnl+} + SSL   & \textbf{94.03} & \textbf{90.78}  & 65.52 & 63.81 & 90.57 & 88.69 & 56.90 & 57.01\\
        & Cluster + UNICON$^{k}$~\cite{karim2022unicon, wang2023lnl+} + SSL  & 92.18   & 80.69  & \textbf{75.52}  & \textbf{73.44} & \textbf{91.63} & \textbf{90.36} & \textbf{65.69} & \textbf{66.00}\\
        \hline
        \textbf{(g)} & Oracle training   & 93.34  & 92.81 & 74.42 & 73.73 & 90.85 & 89.34 & 55.85 & 64.15\\
        \hline
    \end{tabular}
    \caption{\textbf{Synthesized noise results.} \textbf{(a)} Standard training without LNL methods. \textbf{(b)} LNL regularization methods. \textbf{(c)} LNL noise modeling methods. \textbf{(d)} LNL noise detection methods. \textbf{(e)} LNL noise modeling methods combined with LNL+K methods (only train with clean samples). \textbf{(f)} LNL noise modeling methods combined with LNL+K methods and use semi-supervised learning (\textit{SSL}) on noisy samples. \textbf{(g)} Training exclusively with the clean samples within the corrupted datasets. Note that this is an ideal scenario, where the clean samples are given; We highlight the best test accuracy across all the methods (except \textit{(g) Oracle training}) in bold. Combining all three components in our \ourmethodspace framework, methods in \textit{(f)}, achieve the best accuracy in seven out of eight settings, highlighting the importance of utilizing all three components. Interestingly, using \ourmethodspace framework even achieves better performance compared to \textit{Oracle training} in low-noise ratio scenarios. See Section~\ref{sec:results} for discussion.}  
    \label{tab:cifar_results}
\end{table*}

\textbf{Synthesized noise.}
\label{result:combination}
We experiment with different methods for each component under \ourmethodspace framework and report accuracy scores under different noise settings in Table~\ref{tab:cifar_results}. Overall, the combination methods under our \ourmethodspace framework outperform the individual ones, and the advantage is more significant at lower noise ratios. Incorporating semi-supervised learning increases the performance by 4\% on average and even achieves higher accuracy scores than \textit{Oracle training} in six noise settings. For example, at a dominant noise level of 20\%, our method outperforms \textit{Oracle training} by 10\%.

In Fig.~\ref{fig:parallel_charts}, we show the effect of integrating different model choices for the three components under \ourmethodspace framework. For single-component baseline methods, ``None" is placed on the axis.  For example, applying DualT~\cite{yao2020dual} alone is represented as ``DualT-None-None-None". Each line represents a specific method combination, with the optimal one evident by tracing from the top of the accuracy axis. The green line distribution on each axis indicates the significance of that method. In general, not including one or more components (denoted as ``None") results in reduced accuracy, indicating that the integration of these components is essential for effectively managing noisy data. Interestingly, employing Semi-supervised training (SSL) consistently delivers the highest performance among all the combinations. In addition, a comparison between Fig.~\ref{fig:parallel_charts}a and Fig.~\ref{fig:parallel_charts}c reveals that, with 20\% pairwise noise, estimation methods contribute more evenly in the former than in the latter, underscoring the increasing importance of noise estimation in more complex datasets.

\label{result:F1-accuracy}
Fig.~\ref{Fig:precision_recall_f1_acc} highlights the correlation between sample selection and accuracy, detailed scores can be found in the supplementary material. Points spread horizontally suggest a minor influence on the x-axis score. The plot reveals that in precision-accuracy, more densely filled points spread horizontally, indicating precision's reduced significance in high-noise settings. Conversely, the upward-right clusters, particularly evident in circles representing the simpler CIFAR-10 dataset, exhibit a stronger correlation between recall/F1 and accuracy. This insight suggests that in low-noise settings, particularly in datasets with fewer classes, recall in sample selection holds greater importance than precision. This underscores why semi-supervised learning can significantly enhance performance in such scenarios. Additionally, vertical line distributions of densely filled points imply that even with identical F1/Recall scores, variations in accuracy exist. This observation suggests that in high noise settings, the quality alongside the quantity of selected samples plays a pivotal role in performance differences.

\smallskip
\label{result:F1}
\noindent \textbf{Real-world natural noise.} Table \ref{tb:realdataset_results} reports results for the real-world noisy datasets. In the cell dataset CHAMMI-CP~\cite{CHAMMI}, the considerable similarity between ``weak" treatments and the ``control" group introduces notably high noise ratios, profoundly impacting the accuracy score. while most single-component baseline methods either display no significant improvement or even a reduction in performance compared to standard training, our COMBO structure method achieves enhancements of up to 8\%. For Animal-10N dataset~\cite{song2019selfie}, combining noise modeling methods improves 9\% for FINE~\cite{kim2021fine} and 3\% for UNICON~\cite{karim2022unicon}.

\begin{table}
  \centering
  \resizebox{\columnwidth}{!}{
      \setlength{\tabcolsep}{1pt}
  \begin{tabular}{lcc}
    \toprule
     & CHAMMI-CP & Animal10N \\
    \midrule
    Standard Training  & 71.54$\pm$\footnotesize{0.45} & 80.32$\pm$\footnotesize{0.20}\\
    DualT~\citep{yao2020dual}  & 70.73$\pm$\footnotesize{0.17}  & 81.14$\pm$\footnotesize{0.28}\\
    TVD~\cite{zhang2021learningtvd}  &  72.91$\pm$\footnotesize{0.61}  & 88.22$\pm$\footnotesize{0.66}\\
    ROBOT~\cite{yong2022holistic}  & 73.37$\pm$\footnotesize{0.54}  &  89.26$\pm$\footnotesize{0.23}\\
    FINE~\citep{kim2021fine} & 67.27$\pm$\footnotesize{0.82} & 81.15$\pm$\footnotesize{0.11}\\
    UNICON~\citep{karim2022unicon}  & 71.45$\pm$\footnotesize{0.03}  & 87.76$\pm$\footnotesize{0.06}\\
 Cluster + FINE$^{k}$~\cite{kim2021fine, wang2023lnl+} + SSL  & 72.21$\pm$\footnotesize{0.72}  & 90.04$\pm$\footnotesize{0.27}\\
    Cluster + UNICON$^{k}$~\cite{karim2022unicon, wang2023lnl+} + SSL & \textbf{79.17$\pm$\footnotesize{0.23}}  & \textbf{90.21$\pm$\footnotesize{0.21}}\\
    \bottomrule
  \end{tabular}
}
    \caption{\textbf{Real-world noisy data results.} By combining all three components in \ourmethodspace framework, \textit{Cluster + UNICON$^{k}$ + SSL}  performs the best on both datasets. Specifically, it outperforms \textit{Standard Training} by a margin of 8-10\%, while consistently outperforming other baseline methods. See Section~\ref{sec:results} for discussion. The best test accuracy is marked in bold. 
  }
  \label{tb:realdataset_results}
\end{table}

%% file: sec/6_conclusion.tex
\section{Conclusion}
This paper introduces \ourmethod, a unified framework that connects noise modeling and detection methods through a noise source identification module. 
By integrating these traditionally separate strategies, our framework offers a comprehensive solution to the challenges of learning with noisy data.
Our extensive experiments, highlighting up to 10\% enhancement on synthesized datasets and 3\%-5\% on real-world datasets, emphasize the significance of leveraging all components of our framework. Additionally, our insights into the correlation between F1 scores in clean sample selection and classification accuracy in various noise settings open intriguing avenues for future research, aiding the design of LNL methods tailored to specific noise settings.

\smallbreak
\noindent\textbf{Acknowledgments and Disclosure of Funding.} This study was supported, in part, by National Science Foundation NSF-DBI awards 2134696.
 

%% file: sec/X_suppl.tex
\clearpage
\setcounter{page}{1}
\maketitlesupplementary


\section{Method Algorithms}
In Section~\ref{sec:COMBO_methods}, we present an overview of the training process using \ourmethod, and present the updates to the training dataset as shown in Figure~\ref{Fig:Overall}. Algorithm~\ref{alg:overall_algo} provides a detailed view of the training procedure.

In Section~\ref{sec:estimation methods}, we present our noise estimation technique using the growing cluster method. We provide a detailed view of this approach in Algorithm~\ref{alg:cluster_algo}.

\begin{algorithm}
\caption{\ourmethodspace Framework}\label{alg:overall_algo}
\Input{Images $X=\{x_i\}_{i=1}^n$,\\
Noisy labels $\widetilde{Y}=\{\widetilde{y_i}\}_{i=1}^n$,\\
Total number of epochs $E$,\\
Number of warm-up epochs $E_w$,\\
Data loader update frequency $E_u$,\\
Noise source identification $\mathrm{NSI}$,\\
Estimation method $\mathrm{EM}$,\\
Detection method $\mathrm{DM}$,\\
Noise training method $\mathrm{TM}_{\mathrm{noise}}$,\\
Standard training method $\mathrm{TM}_\mathrm{st}$,\\
Model parameters $\theta$}
\For{$e\gets1$ \KwTo $E$}{
    \tcp{Standard training for the first $E_w$ warm-up epochs}
    \uIf{ $e \leq E_w$}{
        $\theta = \mathrm{TM}_\mathrm{st}\left(X, \widetilde{Y}, \theta\right)$\\
        $\mathtt{continue}$\\
    }
    \tcp{After warming up, update training set every $E_u$ epochs}
    \uElseIf{$e == E_w+1$ \text{or} $e \% E_u == 1$}{
        \tcp{Noise modeling with estimation method $\mathrm{EM}$ to get a transition/confusion matrix $M$}
        $M$ = $\mathrm{EM}(X, \widetilde{Y})$\\
        \tcp{Noise source identification to get a dictionary $\mathrm{NS}$}
        $\mathrm{NS}$ = $\mathrm{NSI}(M)$\\
        \tcp{Update the training set with noisy label detection method $\mathrm{DM}$ and noise sources $\mathrm{NS}$}
        $X', \widetilde{Y}'$ = $\mathrm{DM}(X$, $\widetilde{Y}$, $\mathrm{NS})$\\
    }
    \tcp{Training with the updated dataset $(X', \widetilde{Y}')$ and the noise training method $\mathrm{TM}_{\mathrm{noise}}$}
    $\theta$ = $\mathrm{TM}_{\mathrm{noise}}(X'$, $\widetilde{Y}', \theta)$\\
}
\Output{Final trained model parameters $\theta$}
\end{algorithm}

\begin{algorithm}
\caption{Growth Cluster Algorithm}\label{alg:cluster_algo}
\Input{Images $X=\{x_i\}_{i=1}^n$,\\
Noisy labels $\widetilde{Y}=\{\widetilde{y_i}\}_{i=1}^n$,\\
Features $f(X)=\{f(x_i)\}_{i=1}^n$,\\ 
Sample predictions \rightline{$\hat{Y}= \{\hat{y_i} = \arg\max_{j} p(x_i, j)_{j=1}^K\}_{i=1}^n$,}\\
Predicted probabilities for the predictions \rightline{$P(X, \hat{Y})=\{\{p(x_i, \hat{y_i})\}_{i=1}^n)\}$,}\\
Distance metric $\mathrm{dist}$,\\
Anchor selection percentile threshold $t$}
\tcp{Cluster anchor points selection}
\For{$k\gets1$ \KwTo $K$}{ 
    Initialize set $C_{k}^{init}$  by Eq.~\ref{eq:cluster_s1} using \\ $\qquad$  $f(X), \widetilde{Y}, \hat{Y}, P(X, \hat{Y}), t$ \\
    \uIf{$\text{len}(C_{k}^{init}) == 0$}
    {Initialize set $C_{k}^{init}$ by Eq.~\ref{eq:cluster_s2} using\\ $\qquad$ $f(X), \widetilde{Y}, \hat{Y}, P(X, \hat{Y}), t$\\
    } 
    Compute centroid $C_k^{\text{\textit{*}}}$ for each cluster $k$ by Eq.~\ref{eq:cluster_centroid}
}
Order the non-anchor samples $X_{\mathrm{rest}}$ in descending order of $P(X, \hat{Y})$\\
\tcp{Add the rest of data points to the clusters}
\For{$i\gets1$ \KwTo $\text{len}(X_{\mathrm{rest}})$}{ 
    Find the nearest cluster to data point $x_i$\\ $\qquad$ $k=\arg \min_j{\mathrm{dist}(C^*_{j}, x_i)}$\\
    Add $x_i$ to the selected cluster $k$\\
    Update the cluster centroid $C_k^{\text{\textit{*}}}$ by Eq.~\ref{eq:cluster_centroid}
}
\Output{Clusters $\{C_k^{\text{\textit{*}}}\}_1^K$}
\end{algorithm}

\section{Experimental settings}
\subsection{Synthesized Noise}
We follow the noise generation procedure outlined in LNL+K~\cite{wang2023lnl+} for both pairwise and dominant noise in CIFAR-10/CIFAR-100 datasets. 
In a dominant noise setting, the dataset incorporates both ``recessive" and ``dominant" classes, indicating a potential for mislabeling samples from ``recessive" classes as ``dominant" ones. For instance, in CIFAR-10, the last five categories represent ``recessive" classes, while the first five signify ``dominant" classes. Samples from categories 6-10 might be mislabeled as categories 1-5. This division between ``recessive" and ``dominant" classes in CIFAR-100 also occurs between the last and first 50 categories. The varying numbers of samples are mixed for different noise ratios, ensuring that the dataset remains balanced even after mislabeling.
It is important to note that while the 40\% noise setting in dominant noise has higher noise ratios than the 20\% setting, there are differences in dataset composition and total number of training samples between these two noise ratio settings. Therefore, direct comparisons of classification accuracy are not appropriate.
See Tab.~\ref{tb:dominant-noise} for the noisy data composition details. 
\begin{table}[h]
    \setlength{\tabcolsep}{4pt}
    \centering
    \begin{tabular}{lccc}
        \toprule
        \multicolumn{4}{l}{CIFAR-10 Dominant Noise}\\
        \hline
        Noise ratio & 0.2 & 0.4\\
        Dominant class & 2000 & 1800\\
        Recessive class & 3000 & 4200\\
        \bottomrule
        \multicolumn{4}{l}{CIFAR-100 Dominant Noise}\\
        \hline
        Noise ratio & 0.2 & 0.4\\
        Dominant class & 200 & 180\\
        Recessive class & 300 & 420\\
    \bottomrule
    \end{tabular}
    \caption{Sample composition for CIFAR-10/CIFAR-100 dominant noise.}
    \label{tb:dominant-noise}
\end{table}

\subsection{Backbone Networks}
We used ResNet architectures~\cite {he2016deep} for all approaches, in particular a pre-trained ResNet34 on CIFAR-10/CIFAR-100 and ResNet50 on Animal10N datasets.
All methods used ConvNeXt~\citep{liu2022convnet} for CHAMMI-CP. To accommodate the 5 channel images in CHAMMI-CP, we replaced the first convolutional layer in the network to support the additional image dimensions.
\smallskip

\noindent\textbf{UNICON in \ourmethod.} UNICON~\cite{karim2022unicon} employs a dual-network architecture and includes a Semi-Supervised Learning (SSL) module. To ensure fair comparisons, all baseline UNICON results are obtained by training a single network.
To delineate the roles of noise detection and training methods within the \ourmethodspace framework, UNICON is divided into two distinct parts. When referring to UNICON in the context of \ourmethodspace alongside SSL, it specifically denotes the utilization of only the noise detection part of UNICON (\eg, methods categorized under group (e) in the results tables).

\subsection{Hyperparameters}
For a fair comparison, we use the same hyperparameter settings as in prior work~\cite{kim2021fine, karim2022unicon} for the CIFAR-10/CIFAR-100~\cite{krizhevsky2009cifar} datasets. All the experiments use the same batch size of 128. For CIFAR-10/CIFAR-100~\cite{krizhevsky2009cifar} datasets, the dataloader is updated every 10 epochs after warmup, whereas CHAMMI-CP and Animal10N datasets get updated every 5 epochs. All the other hyperparameters for each dataset are summarized in Table \ref{tab:hyperparam}.

\begin{table}
    \centering
    \begin{tabular}{lccc}
    \toprule
      & Learning rate & Warm-up epochs\\
      \hline
     CIFAR-10 & $1e\text{-}2$ & 20\\
     CIFAR-100 & $1e\text{-}2$ & 40\\
     CHAMMI-CP & $2e\text{-}4$ & 5\\
     Animal10N & $5e\text{-}3$ & 3\\
     \bottomrule
    \end{tabular}
    \caption{Hyperparameters for each dataset.}
    \label{tab:hyperparam}
\end{table}

\section{Additional Results}
\subsection{Results on Clothing1M Dataset}
We conducted experiments on the Clothing1M dataset by utilizing 256,000 randomly selected noisy training images as the training set. Our backbone network was ResNet50~\cite{he2016deep}. All experiments were trained using two GPUs with a batch size of 64, a warm-up period of 1 epoch, and a total of 15 epochs. The initial learning rate was set at 0.02 and decreased by a factor of 0.1 every 5 epochs.

As the results are shown in Tab.~\ref{tab:clothing}, combining noise modeling methods with FINE~\cite{kim2021fine} boosts the performance by 1-2\%. Similarly, using COMBO framework with UNICON~\cite{karim2022unicon} results in a 1\% accuracy boost.
\begin{table}
    \centering
    \begin{tabular}{lcc}
    \toprule
      & Classification accuracy \\
      \hline
     Standard Training & 69.45\footnotesize{$\pm$0.31}\\
     ROBOT~\cite{yong2022holistic} & 69.59\footnotesize{$\pm$0.24}\\
     TVD~\cite{zhang2021learningtvd} & 71.75\footnotesize{$\pm$0.30}\\
     FINE~\cite{kim2021fine} & 70.19\footnotesize{$\pm$0.61}\\
     ROBOT~\cite{yong2022holistic} + FINE$^{+k}$~\cite{kim2021fine, wang2023lnl+} & 71.35\footnotesize{$\pm$0.01}\\
     TVD~\cite{zhang2021learningtvd} + FINE$^{+k}$~\cite{kim2021fine, wang2023lnl+} & \textbf{72.24\footnotesize{$\pm$0.03}}\\
     \hline
     UNICON~\cite{karim2022unicon} & 70.28\footnotesize{$\pm$0.16}\\
     Cluster + UNICON$^{+k}$~\cite{karim2022unicon, wang2023lnl+} + SSL & 71.29\footnotesize{$\pm$0.06} \\
     \bottomrule
    \end{tabular}
    \caption{\textbf{Clothing1M results.} Bold marks the best performance. Methods of \ourmethodspace framework show 1-2\% accuracy gain compared to the single baseline methods.}
    \label{tab:clothing}
\end{table}

\subsection{Detailed Synthesized Noise Results}
We present the relationship between sample detection precision/recall/F1 and classification accuracy in Fig.~\ref{Fig:precision_recall_f1_acc}. For detailed result values along with standard deviation across three separate runs using distinct random seeds, please refer to Tab.~\ref{tab:cifar10_asym_20_results}-Tab.~\ref{tab:cifar100_ctl_40_results}. 

\begin{figure*}[t]
\centering
\begin{subfigure}[t]{\columnwidth}
    \centering
    \includegraphics[width=0.92\textwidth]{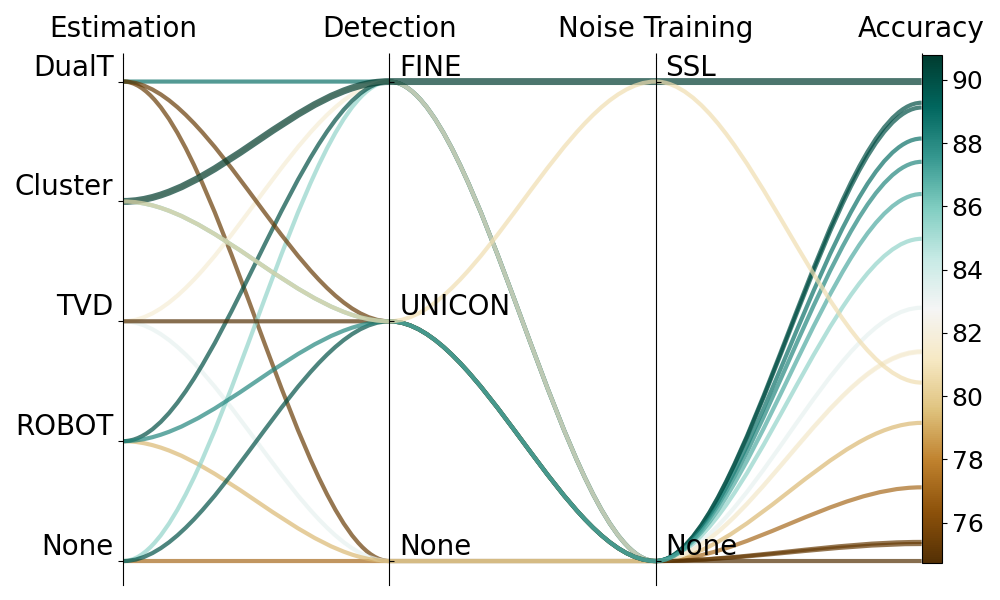}
    \caption{CIFAR-10 40\% pairwise noise}
    \label{fig:cifar10_asym04}
\end{subfigure}
\begin{subfigure}[t]{\columnwidth}
    \centering
    \includegraphics[width=0.92\textwidth]{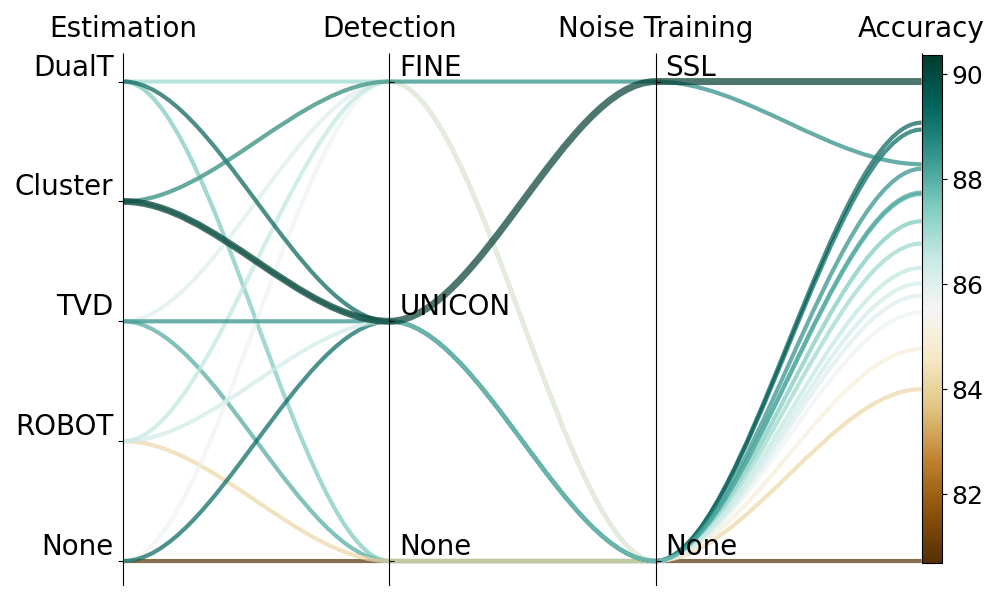}
    \caption{CIFAR-10 40\% dominant noise}
    \label{fig:cifar10_dominant04}
\end{subfigure}
\begin{subfigure}[t]{\columnwidth}
    \centering
    \includegraphics[width=0.92\textwidth]{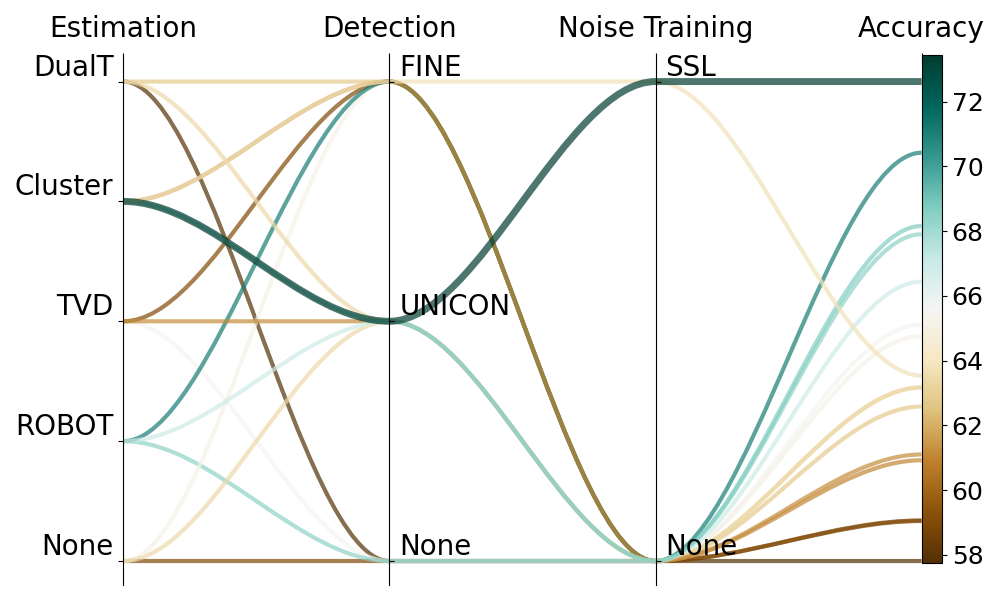}
    \caption{CIFAR-100 40\% pairwise noise}
    \label{fig:cifar100_asym04}
\end{subfigure}
\begin{subfigure}[t]{\columnwidth}
    \centering
    \includegraphics[width=0.92\textwidth]{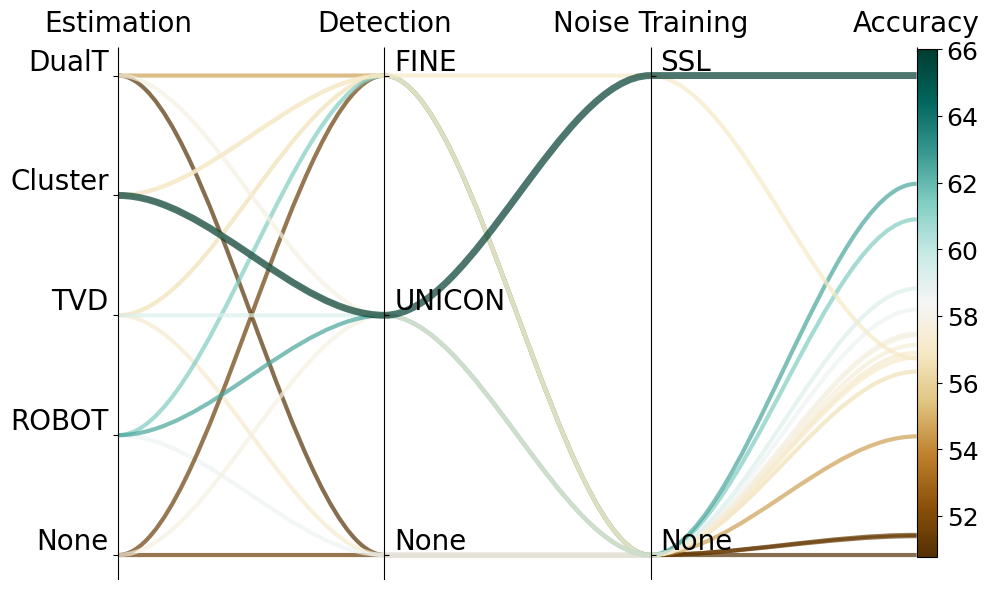}
    \caption{CIFAR-100 40\% dominant noise}
    \label{fig:cifar100_dominant04}
\end{subfigure}
\caption{\textbf{Component interactions in \ourmethodspace framework under 40\% noise.} The effect of combining various methods within the three components of \ourmethod: \textit{noise estimation}, \textit{noise detection}, and \textit{noise training}. Each line in the plot corresponds to a different combination of these components. Generally, omitting one or more components (indicated by \textit{None}) results in reduced accuracy, indicating the importance of integrating these elements to effectively manage noisy data.}
\label{fig:parallel_charts_suppl}
\end{figure*}

Fig.~\ref{fig:parallel_charts} presents the component interaction in \ourmethod framework under 20\% noise ratios. See Fig.~\ref{fig:parallel_charts_suppl} for similar plots of 40\% noise ratio settings.

\begin{table*}[t]
    \centering
    \setlength{\tabcolsep}{2.5pt}
    \begin{tabular}{rlcccccccc}

\toprule
    & & \multicolumn{4}{c}{CIFAR-10 20\% Pairwise Noise}\\
    \cmidrule(lr){3-6}
    \cmidrule(lr){7-10}
     & & \multicolumn{3}{c}{Noise Detection} & Accuracy\\
     \cmidrule(lr){3-5}
     & & Precision & Recall & F1 &\\
     \hline
    \textbf{(a)} & Standard training &  89.12\footnotesize{$\pm$0.00} & 100.00\footnotesize{$\pm$0.00} & 94.25\footnotesize{$\pm$0.00} & 86.12\footnotesize{$\pm$0.09}\\
        \hline
        \textbf{(b)} & SOP~\cite{liu2022robust} & - & - & - & 92.85\footnotesize{$\pm$0.49}\\
        \hline
        \textbf{(c)} 
        & DualT~\cite{yao2020dual} & - & - & - & 90.28\footnotesize{$\pm$0.10}\\
        & TVD~\cite{zhang2021learningtvd}  & - & - & - & 91.50\footnotesize{$\pm$0.02}\\
        & ROBOT~\cite{yong2022holistic}  & - & - & - & 89.57\footnotesize{$\pm$0.06}\\
        \hline 
        \textbf{(d)} 
        & FINE~\cite{kim2021fine} & 99.66\footnotesize{$\pm$0.01} & 70.23\footnotesize{$\pm$3.90} & 82.40\footnotesize{$\pm$2.55} & 89.07\footnotesize{$\pm$0.03}\\
        & UNICON~\cite{karim2022unicon} & 98.21\footnotesize{$\pm$0.06} & 72.71\footnotesize{$\pm$1.21} & 83.56\footnotesize{$\pm$0.11} & 90.31\footnotesize{$\pm$0.04}\\
        \hline
        \textbf{(e)} 
        & DualT~\cite{yao2020dual} + FINE$^{k}$~\cite{kim2021fine, wang2023lnl+} & \textbf{99.77\footnotesize{$\pm$0.01}} & 81.65\footnotesize{$\pm$2.47} & 89.81\footnotesize{$\pm$1.50} & 89.89\footnotesize{$\pm$0.17}\\
        & TVD~\cite{zhang2021learningtvd} + FINE$^{k}$~\cite{kim2021fine, wang2023lnl+} & 97.15\footnotesize{$\pm$0.06} & 84.90\footnotesize{$\pm$2.21} & 90.61\footnotesize{$\pm$1.24} & 90.32\footnotesize{$\pm$0.08}\\
        & ROBOT~\cite{yong2022holistic} + FINE$^{k}$~\cite{kim2021fine, wang2023lnl+} & 99.70\footnotesize{$\pm$0.03} & 85.68\footnotesize{$\pm$0.39} & 92.16\footnotesize{$\pm$0.24} & 90.91\footnotesize{$\pm$0.38}\\
        & Cluster + FINE$^{k}$~\cite{kim2021fine, wang2023lnl+} & 98.30\footnotesize{$\pm$1.50}  & 82.91\footnotesize{$\pm$3.91} & 89.95\footnotesize{$\pm$2.89}  & 88.44\footnotesize{$\pm$2.18} \\
        & DualT~\cite{yao2020dual} + UNICON$^{k}$~\cite{karim2022unicon, wang2023lnl+} & 98.36\footnotesize{$\pm$0.11} & \textbf{95.58\footnotesize{$\pm$0.45}} & 96.95\footnotesize{$\pm$0.28} & 90.96\footnotesize{$\pm$0.32}\\
        & TVD~\cite{zhang2021learningtvd} + UNICON$^{k}$~\cite{karim2022unicon, wang2023lnl+} & 96.74\footnotesize{$\pm$1.51} & \textbf{95.50\footnotesize{$\pm$1.68}} & 96.12\footnotesize{$\pm$1.60} & 91.31\footnotesize{$\pm$0.62}\\
        & ROBOT~\cite{yong2022holistic} + UNICON$^{k}$~\cite{karim2022unicon, wang2023lnl+} & 98.41\footnotesize{$\pm$0.12}  & \textbf{96.32\footnotesize{$\pm$0.40}} & \textbf{97.35\footnotesize{$\pm$0.27}} & 91.57\footnotesize{$\pm$0.08}\\
        & Cluster + UNICON$^{k}$~\cite{karim2022unicon, wang2023lnl+} & 89.01\footnotesize{$\pm$0.08}  & 89.23\footnotesize{$\pm$0.13}  & 89.12\footnotesize{$\pm$0.10}  & 91.63\footnotesize{$\pm$0.21}\\
     \hline
        \textbf{(f)} 
        & Cluster + FINE$^{k}$~\cite{kim2021fine, wang2023lnl+} + SSL & \textbf{99.76\footnotesize{$\pm$0.03}}  & 93.21\footnotesize{$\pm$1.20} & 96.37\footnotesize{$\pm$0.63} & \textbf{94.03\footnotesize{$\pm$0.28}}\\
        & Cluster + UNICON$^{k}$~\cite{karim2022unicon, wang2023lnl+} + SSL & 88.91\footnotesize{$\pm$0.15} & 89.42\footnotesize{$\pm$0.07} & 89.16\footnotesize{$\pm$0.10} & 92.18\footnotesize{$\pm$0.31}\\
        \hline
        \textbf{(g)} & Oracle training &  100.00\footnotesize{$\pm$0.00} & 100.00\footnotesize{$\pm$0.00} & 100.00\footnotesize{$\pm$0.00} & 93.34\footnotesize{$\pm$0.03}\\
        \bottomrule
\end{tabular}
    \caption{\textbf{CIFAR-10 20\% Pairwise Noise.} Bold marks the best performances. Notably, integrating noise modeling methods elevates the sample selection recall rate by up to 25\%. Moreover, the leading model within the COMBO framework outperforms Oracle training by nearly 1\%. See Sec.~\ref{sec:results} for more discussion.}
    \label{tab:cifar10_asym_20_results}
\end{table*}

\begin{table*}[t]
    \centering
    \setlength{\tabcolsep}{2.5pt}
    \begin{tabular}{rlcccccccc}

\toprule
    & & \multicolumn{4}{c}{CIFAR-10 40\% Pairwise Noise}\\
    \cmidrule(lr){3-6}
     & & \multicolumn{3}{c}{Noise Detection} &  Accuracy\\
     \cmidrule(lr){3-5}
     & & Precision & Recall & F1 &\\
     \hline
    \textbf{(a)} 
        & Standard training & 80.96\footnotesize{$\pm$0.00} & 100.00\footnotesize{$\pm$0.00} & 89.48\footnotesize{$\pm$0.00} & 77.18\footnotesize{$\pm$0.27}\\
        \hline
        \textbf{(b)} 
        & SOP~\cite{liu2022robust} & - & - & - & 89.93\footnotesize{$\pm$0.25}\\
        \hline
        \textbf{(c)} 
        & DualT~\cite{yao2020dual} & - & - & - & 75.35\footnotesize{$\pm$0.03}\\
        & TVD~\cite{zhang2021learningtvd} & - & - & - & 83.20\footnotesize{$\pm$0.04}\\
        & ROBOT~\cite{yong2022holistic} & - & - & - & 79.34\footnotesize{$\pm$0.19} \\
        \hline 
        \textbf{(d)} 
        & FINE~\cite{kim2021fine} & 91.86\footnotesize{$\pm$0.34} & 71.56\footnotesize{$\pm$1.15} & 80.45\footnotesize{$\pm$0.80} & 85.51\footnotesize{$\pm$0.18}\\
        & UNICON~\cite{karim2022unicon} & 94.71\footnotesize{$\pm$2.36} & 50.11\footnotesize{$\pm$0.97} & 65.54\footnotesize{$\pm$0.37} & 89.91\footnotesize{$\pm$0.12}\\
        \hline
        \textbf{(e)} & DualT~\cite{yao2020dual} + FINE$^{k}$~\cite{kim2021fine, wang2023lnl+} & 97.72\footnotesize{$\pm$0.44} & 82.68\footnotesize{$\pm$3.48} & 89.57\footnotesize{$\pm$2.16} & 88.87\footnotesize{$\pm$0.49}\\
        & TVD~\cite{zhang2021learningtvd} + FINE$^{k}$~\cite{kim2021fine, wang2023lnl+} & 83.85\footnotesize{$\pm$4.24} & 89.51\footnotesize{$\pm$3.59} & 86.59\footnotesize{$\pm$3.92} & 81.71\footnotesize{$\pm$0.24}\\
        & ROBOT~\cite{yong2022holistic} + FINE$^{k}$~\cite{kim2021fine, wang2023lnl+} & \textbf{98.55\footnotesize{$\pm$0.00}}  & 89.02\footnotesize{$\pm$0.02} & 93.54\footnotesize{$\pm$0.01} & 90.07\footnotesize{$\pm$0.06}\\
        & Cluster + FINE$^{k}$~\cite{kim2021fine, wang2023lnl+} & 87.58\footnotesize{$\pm$0.18} & 80.40\footnotesize{$\pm$0.16} & 83.84\footnotesize{$\pm$0.01} & 81.74\footnotesize{$\pm$0.13}\\
        & DualT~\cite{yao2020dual} + UNICON$^{k}$~\cite{karim2022unicon, wang2023lnl+} & 87.59\footnotesize{$\pm$4.11} & 84.44\footnotesize{$\pm$1.34} & 85.99\footnotesize{$\pm$1.36} & 75.28\footnotesize{$\pm$0.99}\\
        & TVD~\cite{zhang2021learningtvd} + UNICON$^{k}$~\cite{karim2022unicon, wang2023lnl+} & 81.83\footnotesize{$\pm$2.99} & \textbf{96.34\footnotesize{$\pm$0.98}} & 88.50\footnotesize{$\pm$1.44} & 74.71\footnotesize{$\pm$0.30}\\
        & ROBOT~\cite{yong2022holistic} + UNICON$^{k}$~\cite{karim2022unicon, wang2023lnl+} & 94.49\footnotesize{$\pm$0.00} & 95.05\footnotesize{$\pm$0.02} & \textbf{94.77\footnotesize{$\pm$0.01}} &88.09\footnotesize{$\pm$0.12} \\
        & Cluster + UNICON$^{k}$~\cite{karim2022unicon, wang2023lnl+} & 78.07\footnotesize{$\pm$2.18} & 85.25\footnotesize{$\pm$0.79} & 81.50\footnotesize{$\pm$1.57} & 87.01\footnotesize{$\pm$0.98}\\
     \hline
        \textbf{(f)} 
        & Cluster + FINE$^{k}$~\cite{kim2021fine, wang2023lnl+} + SSL & 93.09\footnotesize{$\pm$2.56} & 90.97\footnotesize{$\pm$0.62} & 92.02\footnotesize{$\pm$1.66} & \textbf{90.78\footnotesize{$\pm$0.59}}\\
        & Cluster + UNICON$^{k}$~\cite{karim2022unicon, wang2023lnl+} + SSL & 76.80\footnotesize{$\pm$0.05} & 90.84\footnotesize{$\pm$0.69} & 83.23\footnotesize{$\pm$0.25} & 80.69\footnotesize{$\pm$0.88}\\
        \hline
        \textbf{(g)} 
        & Oracle training & 100.00\footnotesize{$\pm$0.00} & 100.00\footnotesize{$\pm$0.00} & 100.00\footnotesize{$\pm$0.00} & 92.81\footnotesize{$\pm$0.09} \\
        \bottomrule
\end{tabular}
    \caption{\textbf{CIFAR-10 40\% Pairwise Noise.} Bold marks the best performances. Notably, integrating noise modeling methods elevates the sample selection recall rate by up to 45\% and boosts the classification accuracy by 5\%. See Sec.~\ref{sec:results} for more discussion.}
    \label{tab:cifar10_asym_40_results}
\end{table*}

\begin{table*}[t]
    \centering
    \setlength{\tabcolsep}{2.5pt}
    \begin{tabular}{rlcccccccc}

\toprule
    & & \multicolumn{4}{c}{CIFAR-100 20\% Pairwise Noise}\\
    \cmidrule(lr){3-6}
     & & \multicolumn{3}{c}{Noise Detection} & Accuracy\\
     \cmidrule(lr){3-5}
     & & Precision & Recall & F1 & \\
     \hline
    \textbf{(a)} 
        & Standard training &  95.64\footnotesize{$\pm$0.00} & 100.00\footnotesize{$\pm$0.00} & 97.77\footnotesize{$\pm$0.00} & 62.96\footnotesize{$\pm$0.05}\\
        \hline
        \textbf{(b)} 
        & SOP~\cite{liu2022robust} & - & - & - & 72.60\footnotesize{$\pm$0.70}\\
        \hline
        \textbf{(c)} 
        & DualT~\cite{yao2020dual} & - & - & - & 62.76\footnotesize{$\pm$1.49}\\
        & TVD~\cite{zhang2021learningtvd} & - & - & - & 72.62\footnotesize{$\pm$0.01}\\
        & ROBOT~\cite{yong2022holistic}  & - & - & - & 71.16\footnotesize{$\pm$0.40}\\
        \hline
        \textbf{(d)} 
        & FINE~\cite{kim2021fine} & 99.61\footnotesize{$\pm$0.00} & 46.91\footnotesize{$\pm$1.47} & 63.78\footnotesize{$\pm$1.34} & 65.42\footnotesize{$\pm$0.11}\\
        & UNICON~\cite{karim2022unicon} & 99.62\footnotesize{$\pm$0.02}  & 50.91\footnotesize{$\pm$0.28} & 67.38\footnotesize{$\pm$0.24} & 66.87\footnotesize{$\pm$0.04} \\
        \hline
        \textbf{(e)} 
        & DualT~\cite{yao2020dual} + FINE$^{k}$~\cite{kim2021fine, wang2023lnl+} & 98.70\footnotesize{$\pm$0.77} & 65.47\footnotesize{$\pm$3.73} & 78.72\footnotesize{$\pm$2.60} & 66.36\footnotesize{$\pm$0.92}\\
        & TVD~\cite{zhang2021learningtvd} + FINE$^{k}$~\cite{kim2021fine, wang2023lnl+} & 99.31\footnotesize{$\pm$0.03} & 52.32\footnotesize{$\pm$0.18} & 68.53\footnotesize{$\pm$0.16} & 64.34\footnotesize{$\pm$0.22}\\
        & ROBOT~\cite{yong2022holistic} + FINE$^{k}$~\cite{kim2021fine, wang2023lnl+} & 97.00\footnotesize{$\pm$0.09} & \textbf{98.00\footnotesize{$\pm$0.11}} & \textbf{97.50\footnotesize{$\pm$0.01}} & 73.93\footnotesize{$\pm$0.05}\\
        & Cluster + FINE$^{k}$~\cite{kim2021fine, wang2023lnl+} & 98.31\footnotesize{$\pm$0.10} & 65.69\footnotesize{$\pm$2.45} & 78.76\footnotesize{$\pm$1.87} & 67.60\footnotesize{$\pm$0.16}\\
        & DualT~\cite{yao2020dual} + UNICON$^{k}$~\cite{karim2022unicon, wang2023lnl+} & 98.06\footnotesize{$\pm$0.22} & 80.49\footnotesize{$\pm$0.53} & 88.41\footnotesize{$\pm$0.33} & 67.94\footnotesize{$\pm$0.66}\\
        & TVD~\cite{zhang2021learningtvd} + UNICON$^{k}$~\cite{karim2022unicon, wang2023lnl+} & 98.89\footnotesize{$\pm$0.24} & 78.02\footnotesize{$\pm$0.35} & 87.22\footnotesize{$\pm$0.18} & 67.73\footnotesize{$\pm$0.14}\\
        & ROBOT~\cite{yong2022holistic} + UNICON$^{k}$~\cite{karim2022unicon, wang2023lnl+} & 98.26\footnotesize{$\pm$0.14} & 96.39\footnotesize{$\pm$0.02} & 97.32\footnotesize{$\pm$0.06} & 73.82\footnotesize{$\pm$0.06}\\
        & Cluster + UNICON$^{k}$~\cite{karim2022unicon, wang2023lnl+} & 96.21\footnotesize{$\pm$0.06} & 84.76\footnotesize{$\pm$1.38} & 90.12\footnotesize{$\pm$0.79} & 73.38\footnotesize{$\pm$0.11}\\
        \hline
        \textbf{(f)} 
        & Cluster + FINE$^{k}$~\cite{kim2021fine, wang2023lnl+} + SSL & \textbf{99.71\footnotesize{$\pm$0.04}} & 54.64\footnotesize{$\pm$0.33} & 70.60\footnotesize{$\pm$0.23} & 65.52\footnotesize{$\pm$0.42}\\
        & Cluster + UNICON$^{k}$~\cite{karim2022unicon, wang2023lnl+} + SSL & 96.44\footnotesize{$\pm$0.01} & 83.66\footnotesize{$\pm$0.88} & 89.60\footnotesize{$\pm$0.49} & \textbf{75.52\footnotesize{$\pm$0.11}}\\
        \hline
        \textbf{(g)} 
        & Oracle training &  100.00\footnotesize{$\pm$0.00} & 100.00\footnotesize{$\pm$0.00} & 100.00\footnotesize{$\pm$0.00} & 74.42\footnotesize{$\pm$0.02}\\
        \bottomrule
    \end{tabular}
    \caption{\textbf{CIFAR-100 20\% Pairwise Noise.} Bold marks the best performances. Notably, integrating noise modeling methods elevates the sample selection recall rate by up to 50\%. Moreover, the leading model within the COMBO framework outperforms Oracle training by 1\%. See Sec.~\ref{sec:results} for more discussion.}
    \label{tab:cifar100_asym_20_results}
\end{table*}

\begin{table*}[t]
    \centering
    \setlength{\tabcolsep}{2.5pt}
    \begin{tabular}{rlcccccccc}

\toprule
    & & \multicolumn{4}{c}{CIFAR-100 40\% Pairwise Noise}\\
    \cmidrule(lr){3-6}
     & & \multicolumn{3}{c}{Noise Detection} & Accuracy\\
     \cmidrule(lr){3-5}
     & & Precision & Recall & F1 & \\
     \hline
        \textbf{(a)} 
        & Standard training & 92.24\footnotesize{$\pm$0.00} & 100.00\footnotesize{$\pm$0.00}  & 95.96\footnotesize{$\pm$0.00}  & 59.07\footnotesize{$\pm$0.06} \\
        \hline
        \textbf{(b)} 
        & SOP~\cite{liu2022robust} & - & - & - & 70.58\footnotesize{$\pm$0.30} \\
        \hline
        \textbf{(c)} 
        & DualT~\cite{yao2020dual} & - & - & - & 57.74\footnotesize{$\pm$1.92}  \\
        & TVD~\cite{zhang2021learningtvd} & - & - & - & 65.48\footnotesize{$\pm$0.52} \\
        & ROBOT~\cite{yong2022holistic}  & - & - & - & 68.44\footnotesize{$\pm$0.03}  \\
        \hline
        \textbf{(d)} 
        & FINE~\cite{kim2021fine} & \textbf{97.35\footnotesize{$\pm$0.01}}  & 49.52\footnotesize{$\pm$0.62}  & 65.65\footnotesize{$\pm$0.54}  & 65.11\footnotesize{$\pm$0.11} \\
        & UNICON~\cite{karim2022unicon} & 97.18\footnotesize{$\pm$0.01}  & 49.63\footnotesize{$\pm$0.06}  & 65.70\footnotesize{$\pm$0.06}  & 63.43\footnotesize{$\pm$0.07}  \\
        \hline
        \textbf{(e)} 
        & DualT~\cite{yao2020dual} + FINE$^{k}$~\cite{kim2021fine, wang2023lnl+} & 91.32\footnotesize{$\pm$1.01}  & 65.03\footnotesize{$\pm$0.52}  & 75.97\footnotesize{$\pm$0.11}  & 62.80\footnotesize{$\pm$0.69} \\
        & TVD~\cite{zhang2021learningtvd} + FINE$^{k}$~\cite{kim2021fine, wang2023lnl+} & 86.30\footnotesize{$\pm$0.42}  & 61.36\footnotesize{$\pm$0.08}  & 71.72\footnotesize{$\pm$0.20}  & 59.05\footnotesize{$\pm$0.10} \\
        & ROBOT~\cite{yong2022holistic} + FINE$^{k}$~\cite{kim2021fine, wang2023lnl+} & 92.62\footnotesize{$\pm$0.14}  & \textbf{93.68\footnotesize{$\pm$1.80}}  & \textbf{93.14\footnotesize{$\pm$0.95}}  & 71.11\footnotesize{$\pm$0.12} \\
        & Cluster + FINE$^{k}$~\cite{kim2021fine, wang2023lnl+} & 92.93\footnotesize{$\pm$1.35}  & 58.72\footnotesize{$\pm$1.49}  & 71.97\footnotesize{$\pm$1.52}  & 61.04\footnotesize{$\pm$1.26} \\
        & DualT~\cite{yao2020dual} + UNICON$^{k}$~\cite{karim2022unicon, wang2023lnl+} & 95.25\footnotesize{$\pm$0.46}  & 78.06\footnotesize{$\pm$0.94}  & 85.80\footnotesize{$\pm$0.36}  & 63.42\footnotesize{$\pm$0.57} \\
        & TVD~\cite{zhang2021learningtvd} + UNICON$^{k}$~\cite{karim2022unicon, wang2023lnl+} & 89.50\footnotesize{$\pm$0.96}  & 80.66\footnotesize{$\pm$0.18}  & 84.85\footnotesize{$\pm$0.43}  & 61.23\footnotesize{$\pm$0.48} \\
        & ROBOT~\cite{yong2022holistic} + UNICON$^{k}$~\cite{karim2022unicon, wang2023lnl+} & 95.17\footnotesize{$\pm$0.02}  & 92.39\footnotesize{$\pm$0.15}  & \textbf{93.76\footnotesize{$\pm$0.06}}  & 66.89\footnotesize{$\pm$0.09} \\
        & Cluster + UNICON$^{k}$~\cite{karim2022unicon, wang2023lnl+} & 92.14\footnotesize{$\pm$0.89}  & 82.41\footnotesize{$\pm$0.91}  & 87.00\footnotesize{$\pm$0.90}  & 68.71\footnotesize{$\pm$0.74} \\
        \hline
        \textbf{(f)} 
        & Cluster + FINE$^{k}$~\cite{kim2021fine, wang2023lnl+} + SSL & 95.35\footnotesize{$\pm$0.30}  & 56.82\footnotesize{$\pm$0.59}  & 71.21\footnotesize{$\pm$0.53}  & 63.81\footnotesize{$\pm$0.24} \\
        & Cluster + UNICON$^{k}$~\cite{karim2022unicon, wang2023lnl+} + SSL & 94.29\footnotesize{$\pm$1.41}  & 83.20\footnotesize{$\pm$0.65}  & 88.40\footnotesize{$\pm$0.33}  & \textbf{73.44\footnotesize{$\pm$0.30}} \\
        \hline
        \textbf{(g)} 
        & Oracle training & 100.00\footnotesize{$\pm$0.00}  & 100.00\footnotesize{$\pm$0.00}  & 100.00\footnotesize{$\pm$0.00}  & 73.73\footnotesize{$\pm$0.13} \\
        \bottomrule
\end{tabular}
    \caption{\textbf{CIFAR-100 40\% Pairwise Noise.} Bold marks the best performances. An important observation is the integration of noise modeling methods, which increases the sample selection recall rate by up to 45\% and enhances classification accuracy by 10\%. This leads to a performance level comparable to Oracle within this intermediate noise setting. See Sec.~\ref{sec:results} for more discussion.}
    \label{tab:cifar100_asym_40_results}
\end{table*}

\begin{table*}[t]
    \centering
    \setlength{\tabcolsep}{2.5pt}
    \begin{tabular}{rlcccccccc}

\toprule
    & & \multicolumn{4}{c}{CIFAR-10 20\% Dominant Noise}\\
    \cmidrule(lr){3-6}
     & & \multicolumn{3}{c}{Noise Detection} & Accuracy\\
     \cmidrule(lr){3-5}
     & & Precision & Recall & F1 &\\
     \hline
        \textbf{(a)} 
        & Standard training &  90.00\footnotesize{$\pm$0.00} & 100.00\footnotesize{$\pm$0.00}  & 94.74\footnotesize{$\pm$0.00}  & 85.47\footnotesize{$\pm$0.52} \\
        \hline
        \textbf{(b)} 
        & SOP~\cite{liu2022robust} & - & - & - & 89.86\footnotesize{$\pm$0.40} \\
        \hline
        \textbf{(c)} 
        & DualT~\cite{yao2020dual} & - & - & - & 87.62\footnotesize{$\pm$0.06} \\
        & TVD~\cite{zhang2021learningtvd}  & - & - & - & 88.79\footnotesize{$\pm$0.07} \\
        & ROBOT~\cite{yong2022holistic}  & - & - & - & 86.10\footnotesize{$\pm$0.09} \\
        \hline
        \textbf{(d)} 
        & FINE~\cite{kim2021fine} & \textbf{99.90\footnotesize{$\pm$0.00}}  & 65.58\footnotesize{$\pm$1.64}  & 79.18\footnotesize{$\pm$1.22}  & 86.03\footnotesize{$\pm$0.30} \\
        & UNICON~\cite{karim2022unicon} & \textbf{99.91\footnotesize{$\pm$0.02}}  & 60.03\footnotesize{$\pm$4.51}  & 75.00\footnotesize{$\pm$3.73}  & 90.11\footnotesize{$\pm$0.14} \\
        \hline
        \textbf{(e)} 
        & DualT~\cite{yao2020dual} + FINE$^{k}$~\cite{kim2021fine, wang2023lnl+} & 97.38\footnotesize{$\pm$1.07}  & 87.66\footnotesize{$\pm$7.50}  & 92.27\footnotesize{$\pm$4.45}  & 88.76\footnotesize{$\pm$0.42} \\
        & TVD~\cite{zhang2021learningtvd} + FINE$^{k}$~\cite{kim2021fine, wang2023lnl+} & 95.22\footnotesize{$\pm$0.75}  & 89.77\footnotesize{$\pm$2.21}  & 92.42\footnotesize{$\pm$0.76}  & 87.93\footnotesize{$\pm$0.43} \\
        & ROBOT~\cite{yong2022holistic} + FINE$^{k}$~\cite{kim2021fine, wang2023lnl+} & 95.48\footnotesize{$\pm$0.74}  & 95.92\footnotesize{$\pm$0.58}  & 95.70\footnotesize{$\pm$0.66}  & 88.17\footnotesize{$\pm$0.84} \\
        & Cluster + FINE$^{k}$~\cite{kim2021fine, wang2023lnl+}& 97.46\footnotesize{$\pm$1.78}  & 86.07\footnotesize{$\pm$3.70}  & 91.41\footnotesize{$\pm$1.41}  & 85.46\footnotesize{$\pm$0.21} \\
        & DualT~\cite{yao2020dual} + UNICON$^{k}$~\cite{karim2022unicon, wang2023lnl+} & 96.50\footnotesize{$\pm$0.61}  & 94.50\footnotesize{$\pm$0.23}  & 95.49\footnotesize{$\pm$0.19}  & 89.28\footnotesize{$\pm$0.13} \\
        & TVD~\cite{zhang2021learningtvd} + UNICON$^{k}$~\cite{karim2022unicon, wang2023lnl+} & 96.55\footnotesize{$\pm$0.03}  & \textbf{99.62\footnotesize{$\pm$1.27}}  & 98.06\footnotesize{$\pm$0.66}  & 88.69\footnotesize{$\pm$0.14} \\
        & ROBOT~\cite{yong2022holistic} + UNICON$^{k}$~\cite{karim2022unicon, wang2023lnl+} & 95.56\footnotesize{$\pm$0.01}  & \textbf{99.72\footnotesize{$\pm$0.02}}  & \textbf{97.60\footnotesize{$\pm$0.01}}  & 89.02\footnotesize{$\pm$0.13} \\
        & Cluster + UNICON$^{k}$~\cite{karim2022unicon, wang2023lnl+} & 94.64\footnotesize{$\pm$0.01}  & 95.66\footnotesize{$\pm$0.23}  & 95.15\footnotesize{$\pm$0.11}  & 89.18\footnotesize{$\pm$0.11} \\
     \hline
        \textbf{(f)} 
        & Cluster + FINE$^{k}$~\cite{kim2021fine, wang2023lnl+} + SSL & 98.80\footnotesize{$\pm$1.65}  & 92.53\footnotesize{$\pm$0.20}  & 95.56\footnotesize{$\pm$0.68}  & 90.57\footnotesize{$\pm$0.52} \\
        & Cluster + UNICON$^{k}$~\cite{karim2022unicon, wang2023lnl+} + SSL & 94.60\footnotesize{$\pm$0.05}  & 97.77\footnotesize{$\pm$0.53}  & 96.16\footnotesize{$\pm$0.24}  & \textbf{91.63\footnotesize{$\pm$0.25}} \\
        \hline
        \textbf{(g)} 
        & Oracle training &  100.00\footnotesize{$\pm$0.00}  & 100.00\footnotesize{$\pm$0.00}  & 100.00\footnotesize{$\pm$0.00}  & 90.85\footnotesize{$\pm$0.04} \\
        \bottomrule
    \end{tabular}
     \caption{\textbf{CIFAR-10 20\% Dominant Noise.} Bold marks the best performances. Notably, integrating noise modeling methods elevates the sample selection recall rate by up to 40\%. Moreover, the leading model within the COMBO framework outperforms Oracle training by nearly 1\%. See Sec.~\ref{sec:results} for more discussion.}
    \label{tab:cifar10_ctl_20_results}
\end{table*}

\begin{table*}[t]
    \centering
    \setlength{\tabcolsep}{2.5pt}
    \begin{tabular}{rlcccccccc}

\toprule
    & & \multicolumn{4}{c}{CIFAR-10 40\% Dominant Noise}\\
    \cmidrule(lr){3-6}
     & & \multicolumn{3}{c}{Noise Detection} & Accuracy\\
     \cmidrule(lr){3-5}
     & & Precision & Recall & F1 &\\
     \hline
        \textbf{(a)} 
        & Standard training  & 80.00\footnotesize{$\pm$0.00} & 100.00\footnotesize{$\pm$0.00} & 88.89\footnotesize{$\pm$0.00} & 80.68\footnotesize{$\pm$0.25} \\
        \hline
        \textbf{(b)} 
        & SOP~\cite{liu2022robust} & - & - & - & 89.30\footnotesize{$\pm$0.36}\\
        \hline
        \textbf{(c)} 
        & DualT~\cite{yao2020dual} & - & - & - & 87.54\footnotesize{$\pm$0.04}\\
        & TVD~\cite{zhang2021learningtvd}  & - & - & - & 88.12\footnotesize{$\pm$0.22}\\
        & ROBOT~\cite{yong2022holistic}  & - & - & - & 84.15\footnotesize{$\pm$0.53} \\
        \hline
        \textbf{(d)} 
        & FINE~\cite{kim2021fine} & 99.71\footnotesize{$\pm$0.00} & 68.01\footnotesize{$\pm$3.40} & 80.86\footnotesize{$\pm$2.51} & 85.70\footnotesize{$\pm$0.38}\\
        & UNICON~\cite{karim2022unicon} & \textbf{99.96\footnotesize{$\pm$0.01}} & 66.72\footnotesize{$\pm$0.09} & 80.02\footnotesize{$\pm$0.06} & 89.39\footnotesize{$\pm$0.14}\\
        \hline
        \textbf{(e)} 
        & DualT~\cite{yao2020dual} + FINE$^{k}$~\cite{kim2021fine, wang2023lnl+} & 89.17\footnotesize{$\pm$1.74} & 88.40\footnotesize{$\pm$2.82} & 88.78\footnotesize{$\pm$1.43} & 87.09\footnotesize{$\pm$0.78}\\
        & TVD~\cite{zhang2021learningtvd} + FINE$^{k}$~\cite{kim2021fine, wang2023lnl+} & 84.42\footnotesize{$\pm$5.79} & 77.54\footnotesize{$\pm$9.08} & 80.83\footnotesize{$\pm$7.53} & 86.04\footnotesize{$\pm$0.25}\\
        & ROBOT~\cite{yong2022holistic} + FINE$^{k}$~\cite{kim2021fine, wang2023lnl+} & 85.39\footnotesize{$\pm$0.57}  & 83.75\footnotesize{$\pm$10.00} & 84.57\footnotesize{$\pm$4.37} & 86.60\footnotesize{$\pm$1.11}\\
        & Cluster + FINE$^{k}$~\cite{kim2021fine, wang2023lnl+} & 99.86\footnotesize{$\pm$0.98} & 86.09\footnotesize{$\pm$3.54} & 92.47\footnotesize{$\pm$1.55} & 84.96\footnotesize{$\pm$0.51}\\
        & DualT~\cite{yao2020dual} + UNICON$^{k}$~\cite{karim2022unicon, wang2023lnl+} & 88.46\footnotesize{$\pm$2.24} & 85.22\footnotesize{$\pm$1.41} & 86.81\footnotesize{$\pm$1.85} & 89.53\footnotesize{$\pm$0.01}\\
        & TVD~\cite{zhang2021learningtvd} + UNICON$^{k}$~\cite{karim2022unicon, wang2023lnl+} & 86.66\footnotesize{$\pm$3.43} & 94.23\footnotesize{$\pm$0.38} & 90.28\footnotesize{$\pm$1.95} & 88.60\footnotesize{$\pm$0.02}\\
        & ROBOT~\cite{yong2022holistic} + UNICON$^{k}$~\cite{karim2022unicon, wang2023lnl+} & 81.92\footnotesize{$\pm$5.09} & 80.88\footnotesize{$\pm$5.17}  & 81.40\footnotesize{$\pm$5.12} & 86.29\footnotesize{$\pm$0.55}\\
        & Cluster + UNICON$^{k}$~\cite{karim2022unicon, wang2023lnl+} & 85.49\footnotesize{$\pm$0.04} & 87.29\footnotesize{$\pm$0.16} & 86.38\footnotesize{$\pm$0.09} & 88.09\footnotesize{$\pm$0.18}\\
     \hline
        \textbf{(f)} 
        & Cluster + FINE$^{k}$~\cite{kim2021fine, wang2023lnl+} + SSL & 87.44\footnotesize{$\pm$5.94} & \textbf{99.26\footnotesize{$\pm$1.86}} & \textbf{92.98\footnotesize{$\pm$2.21}} & 88.69\footnotesize{$\pm$1.31}\\
        & Cluster + UNICON$^{k}$~\cite{karim2022unicon, wang2023lnl+} + SSL & 85.33\footnotesize{$\pm$0.58} & 89.28\footnotesize{$\pm$0.80} & 87.26\footnotesize{$\pm$0.09} & \textbf{90.36\footnotesize{$\pm$0.28}}\\
        \hline
        \textbf{(g)} 
        & Oracle training & 100.00\footnotesize{$\pm$0.00} & 100.00\footnotesize{$\pm$0.00} & 100.00\footnotesize{$\pm$0.00} & 89.34\footnotesize{$\pm$0.07} \\
        \bottomrule
    \end{tabular}
     \caption{\textbf{CIFAR-10 40\% Dominant Noise.} Bold marks the best performances. Notably, integrating noise modeling methods elevates the sample selection recall rate by up to 30\%. Moreover, the leading model within the COMBO framework outperforms Oracle training by nearly 1\%. See Sec.~\ref{sec:results} for more discussion.}
    \label{tab:cifar10_ctl_40_results}
\end{table*}

\begin{table*}[t]
    \centering
    \setlength{\tabcolsep}{2.5pt}
    \begin{tabular}{rlcccccccc}

\toprule
    & & \multicolumn{4}{c}{CIFAR-100 20\% Dominant Noise} \\
    \cmidrule(lr){3-6}
     & & \multicolumn{3}{c}{Noise Detection} & Accuracy\\
     \cmidrule(lr){3-5}
     & & Precision & Recall & F1 &\\
     \hline
        \textbf{(a)} 
        & Standard training &  90.00\footnotesize{$\pm$0.00} & 100.00\footnotesize{$\pm$0.00} & 94.74\footnotesize{$\pm$0.00} & 50.37\footnotesize{$\pm$1.67} \\
        \hline
        \textbf{(b)} 
        & SOP~\cite{liu2022robust} & - & - & - & 62.47\footnotesize{$\pm$0.47}\\
        \hline
        \textbf{(b)} 
        & DualT~\cite{yao2020dual}  & - & - & - & 50.80\footnotesize{$\pm$0.55}\\
        & TVD~\cite{zhang2021learningtvd}  & - & - & - & 60.56\footnotesize{$\pm$0.57}\\
        & ROBOT~\cite{yong2022holistic}  & - & - & - & 60.81\footnotesize{$\pm$1.19}\\
        \hline
        \textbf{(c)} 
        & FINE~\cite{kim2021fine} & 99.49\footnotesize{$\pm$0.00} & 60.07\footnotesize{$\pm$0.36} & 74.91\footnotesize{$\pm$0.01} & 53.68\footnotesize{$\pm$0.13}\\
        & UNICON~\cite{karim2022unicon} & \textbf{99.95\footnotesize{$\pm$0.01}} & 50.57\footnotesize{$\pm$0.04} & 67.16\footnotesize{$\pm$0.03} & 60.34\footnotesize{$\pm$0.28}\\
        \hline
        \textbf{(d)} 
        & DualT~\cite{yao2020dual} + FINE$^{k}$~\cite{kim2021fine, wang2023lnl+} & 99.51\footnotesize{$\pm$0.02} & 65.62\footnotesize{$\pm$0.16} & 79.09\footnotesize{$\pm$0.11} & 57.04\footnotesize{$\pm$0.12}\\
        & TVD~\cite{zhang2021learningtvd} + FINE$^{k}$~\cite{kim2021fine, wang2023lnl+} & 99.80\footnotesize{$\pm$0.00} & 52.56\footnotesize{$\pm$0.45} & 68.86\footnotesize{$\pm$0.39} & 58.14\footnotesize{$\pm$0.06}\\
        & ROBOT~\cite{yong2022holistic} + FINE$^{k}$~\cite{kim2021fine, wang2023lnl+} & 94.45\footnotesize{$\pm$0.00} & 99.53\footnotesize{$\pm$0.01} & 96.92\footnotesize{$\pm$0.00} & 61.98\footnotesize{$\pm$0.31}\\
        & Cluster + FINE$^{k}$~\cite{kim2021fine, wang2023lnl+} & 98.55\footnotesize{$\pm$0.16} & 55.16\footnotesize{$\pm$0.25} & 70.73\footnotesize{$\pm$0.23} & 54.41\footnotesize{$\pm$0.16}\\
        & DualT~\cite{yao2020dual} + UNICON$^{k}$~\cite{karim2022unicon, wang2023lnl+} & 99.87\footnotesize{$\pm$0.00} & 74.75\footnotesize{$\pm$0.15} & 85.51\footnotesize{$\pm$0.01} & 57.68\footnotesize{$\pm$0.13}\\
        & TVD~\cite{zhang2021learningtvd} + UNICON$^{k}$~\cite{karim2022unicon, wang2023lnl+} & 99.68\footnotesize{$\pm$0.00} & 72.11\footnotesize{$\pm$0.01} & 83.68\footnotesize{$\pm$0.01} & 57.76\footnotesize{$\pm$0.02}\\
        & ROBOT~\cite{yong2022holistic} + UNICON$^{k}$~\cite{karim2022unicon, wang2023lnl+} & 94.46\footnotesize{$\pm$0.04}  & \textbf{99.76\footnotesize{$\pm$0.02}} & \textbf{97.03}\footnotesize{$\pm$0.01} & 64.04\footnotesize{$\pm$0.45}\\
        & Cluster + UNICON$^{k}$~\cite{karim2022unicon, wang2023lnl+} & 94.56\footnotesize{$\pm$1.31} & 80.89\footnotesize{$\pm$0.50} & 87.19\footnotesize{$\pm$0.88} & 61.57\footnotesize{$\pm$0.60}\\
     \hline
         \textbf{(e)} 
         & Cluster + FINE$^{k}$~\cite{kim2021fine, wang2023lnl+} + SSL & 98.68\footnotesize{$\pm$0.06} & 53.26\footnotesize{$\pm$0.34} & 69.18\footnotesize{$\pm$0.22} & 56.90\footnotesize{$\pm$0.04}\\
        & Cluster + UNICON$^{k}$~\cite{karim2022unicon, wang2023lnl+} + SSL & 94.40\footnotesize{$\pm$0.02} & 78.57\footnotesize{$\pm$0.04} & 85.76\footnotesize{$\pm$0.02} & \textbf{65.69\footnotesize{$\pm$0.18}}\\
        \hline
        \textbf{(f)} 
        & Oracle training &  100.00\footnotesize{$\pm$0.00} & 100.00\footnotesize{$\pm$0.00} & 100.00\footnotesize{$\pm$0.00} & 55.85\footnotesize{$\pm$0.12}\\
        \bottomrule
    \end{tabular}
     \caption{\textbf{CIFAR-100 20\% Dominant Noise.} Bold marks the best performances. Notably, integrating noise modeling methods elevates the sample selection recall rate by up to 50\%. Moreover, the leading model within the COMBO framework outperforms Oracle training by 10\%. See Sec.~\ref{sec:results} for more discussion.}
     \label{tab:cifar100_ctl_20_results}
\end{table*}

\begin{table*}[t]
    \centering
    \setlength{\tabcolsep}{2.5pt}
    \begin{tabular}{rlcccccccc}

\toprule
    & & \multicolumn{4}{c}{CIFAR-100 40\% Dominant Noise} \\
    \cmidrule(lr){3-6}
     & & \multicolumn{3}{c}{Noise Detection} & Accuracy\\
     \cmidrule(lr){3-5}
     & & Precision & Recall & F1 &\\
     \hline
        \textbf{(a)} 
        & Standard training & 80.00\footnotesize{$\pm$0.00} & 100.00\footnotesize{$\pm$0.00} & 88.89\footnotesize{$\pm$0.00} & 51.40\footnotesize{$\pm$0.20}\\
        \hline
        \textbf{(b)} 
        & SOP~\cite{liu2022robust} & - & - & - & 61.33\footnotesize{$\pm$0.58}\\
        \hline
        \textbf{(b)} 
        & DualT~\cite{yao2020dual}  & - & - & - & 50.76\footnotesize{$\pm$0.03}\\
        & TVD~\cite{zhang2021learningtvd}  & - & - & - & 57.19\footnotesize{$\pm$0.01}\\
        & ROBOT~\cite{yong2022holistic}  & - & - & - & 58.56\footnotesize{$\pm$0.10}\\
        \hline
        \textbf{(c)} 
        & FINE~\cite{kim2021fine} & 98.86\footnotesize{$\pm$0.09} & 58.41\footnotesize{$\pm$0.89} & 73.48\footnotesize{$\pm$0.71} & 51.36\footnotesize{$\pm$0.53} \\
        & UNICON~\cite{karim2022unicon} & \textbf{99.93\footnotesize{$\pm$0.01}} & 52.47\footnotesize{$\pm$0.86} & 68.81\footnotesize{$\pm$0.65} & 57.74\footnotesize{$\pm$0.11}\\
        \hline
        \textbf{(d)} 
        & DualT~\cite{yao2020dual} + FINE$^{k}$~\cite{kim2021fine, wang2023lnl+} & 99.49\footnotesize{$\pm$0.06} & 62.03\footnotesize{$\pm$1.78} & 76.41\footnotesize{$\pm$1.34} & 54.53\footnotesize{$\pm$0.10}\\
        & TVD~\cite{zhang2021learningtvd} + FINE$^{k}$~\cite{kim2021fine, wang2023lnl+} & 99.35\footnotesize{$\pm$0.01} & 52.95\footnotesize{$\pm$1.21} & 69.08\footnotesize{$\pm$0.18} & 56.59\footnotesize{$\pm$0.06}\\
        & ROBOT~\cite{yong2022holistic} + FINE$^{k}$~\cite{kim2021fine, wang2023lnl+} & 84.65\footnotesize{$\pm$0.03} & \textbf{99.31\footnotesize{$\pm$0.76}} & \textbf{91.40\footnotesize{$\pm$0.31}} & 61.43\footnotesize{$\pm$0.75}\\
        & Cluster + FINE$^{k}$~\cite{kim2021fine, wang2023lnl+} & 86.57\footnotesize{$\pm$1.57} & 70.89\footnotesize{$\pm$1.53} & 77.95\footnotesize{$\pm$0.37} & 57.04\footnotesize{$\pm$0.13}\\
        & DualT~\cite{yao2020dual} + UNICON$^{k}$~\cite{karim2022unicon, wang2023lnl+} & 99.70\footnotesize{$\pm$0.00} & 76.66\footnotesize{$\pm$0.05} & 86.68\footnotesize{$\pm$0.02} & 57.80\footnotesize{$\pm$0.39}\\
        & TVD~\cite{zhang2021learningtvd} + UNICON$^{k}$~\cite{karim2022unicon, wang2023lnl+} & 96.98\footnotesize{$\pm$0.38} & 77.34\footnotesize{$\pm$0.69} & 86.06\footnotesize{$\pm$0.20} & 59.23\footnotesize{$\pm$0.23}\\
        & ROBOT~\cite{yong2022holistic} + UNICON$^{k}$~\cite{karim2022unicon, wang2023lnl+} & 84.89\footnotesize{$\pm$0.01} & 99.10\footnotesize{$\pm$0.03} & \textbf{91.45\footnotesize{$\pm$0.02}} & 62.56\footnotesize{$\pm$0.12}\\
        & Cluster + UNICON$^{k}$~\cite{karim2022unicon, wang2023lnl+} & 83.36\footnotesize{$\pm$0.07} & 38.62\footnotesize{$\pm$1.31} & 52.78\footnotesize{$\pm$0.61} & 57.44\footnotesize{$\pm$0.98}\\
     \hline
         \textbf{(e)} 
        & Cluster + FINE$^{k}$~\cite{kim2021fine, wang2023lnl+} + SSL & 95.17\footnotesize{$\pm$0.50} & 67.80\footnotesize{$\pm$2.25} & 79.18\footnotesize{$\pm$1.64} & 57.01\footnotesize{$\pm$0.17}\\
        & Cluster + UNICON$^{k}$~\cite{karim2022unicon, wang2023lnl+} + SSL & 84.55\footnotesize{$\pm$0.24} & 75.23\footnotesize{$\pm$0.39} & 79.83\footnotesize{$\pm$0.32}  & \textbf{66.00\footnotesize{$\pm$0.27}}\\
        \hline
        \textbf{(f)} 
        & Oracle training & 100.00\footnotesize{$\pm$0.00} & 100.00\footnotesize{$\pm$0.00} & 100.00\footnotesize{$\pm$0.00} & 64.15\footnotesize{$\pm$0.02}\\
        \bottomrule
    \end{tabular}
     \caption{\textbf{CIFAR-100 40\% Dominant Noise.} Bold marks the best performances. Notably, integrating noise modeling methods elevates the sample selection recall rate by up to 30\%. Moreover, the leading model within the COMBO framework outperforms Oracle training by nearly 2\% in this intermediate noise setting. See Sec.~\ref{sec:results} for more discussion.}
     \label{tab:cifar100_ctl_40_results}
\end{table*}

